\documentclass[preprint,12pt]{elsarticle}

\usepackage{amssymb}
\usepackage{amsmath}
\usepackage{booktabs}
\usepackage{multirow}
\usepackage{soul} 
\usepackage{color, xcolor}
\usepackage{tabularx}
\newcommand{\ie}{\emph{i.e., }}
\newcommand{\eg}{\emph{e.g., }}

\newcommand{\wrt}{\emph{w.r.t. }}

\newcommand{\aka}{\emph{aka. }}

\begin{document}

\begin{frontmatter}



\title{A Comprehensive Evaluation of Large Language Models on Temporal Event Forecasting}




\author[1]{He Chang}
\ead{hechangcuc@cuc.edu.cn}

\affiliation[1]{organization={School of Computer and Cyber Sciences, Communication University of China},
                city={Beijing},
                postcode={100024}, 
                country={China}}

\author[2]{Chenchen Ye}
\ead{ccye@cs.ucla.edu}
\affiliation[2]{organization={Department of Computer Science, University of California, Los Angeles},
                city={Los Angeles},
                postcode={90095}, 
                country={USA}}
                
\author[3]{Zhulin Tao\corref{cor1}}
\affiliation[3]{organization={School of Information and Communication Engineering, Communication University of China},
                city={Beijing},
                postcode={100024}, 
                country={China}}
\ead{taozhulin@gmail.com}

\author[3]{Jie Wu}
\ead{wujie@cuc.edu.cn}

\author[4]{Zhengmao Yang}
\ead{zmyang4671@zju.edu.cn}

\affiliation[4]{organization={College of Computer Science and Technology, Zhejiang University},
                postcode={310058}, 
                city={Hangzhou},
                country={China}}
\author[5]{Yunshan Ma}
\ead{ysma@smu.edu.sg}

\affiliation[5]{organization={School of Computing and Information Systems, Singapore Management University},
                postcode={178902}, 
                country={Singapore}}

\affiliation[6]{organization={School of Computing, National University of Singapore},
                postcode={119077}, 
                country={Singapore}}

\author[1]{Xianglin Huang}

\author[6]{Tat-Seng Chua}
\ead{dcscts@nus.edu.sg}

\cortext[cor1]{Corresponding author}

\begin{abstract}
Recently, Large Language Models (LLMs) have demonstrated great potential in various data mining tasks, such as knowledge question answering, mathematical reasoning, and commonsense reasoning. However, the reasoning capability of LLMs on temporal event forecasting has been under-explored. To systematically investigate their abilities in temporal event forecasting, we conduct a comprehensive evaluation of LLM-based methods for temporal event forecasting. Due to the lack of a high-quality dataset that involves both graph and textual data, we first construct a benchmark dataset, named MidEast-TE-mini. Based on this dataset, we design a series of baseline methods, characterized by various input formats and retrieval augmented generation (RAG) modules. 
From extensive experiments, we find that directly integrating raw texts into the input of LLMs does not enhance zero-shot extrapolation performance. 
In contrast, fine-tuning LLMs with raw texts can significantly improve performance. 
Additionally, LLMs enhanced with retrieval modules can effectively capture temporal relational patterns hidden in historical events. However, issues such as popularity bias and the long-tail problem persist in LLMs, particularly in the retrieval-augmented generation (RAG) method.
These findings not only deepen our understanding of LLM-based event forecasting methods but also highlight several promising research directions. 
We consider that this comprehensive evaluation, along with the identified research opportunities, will significantly contribute to future research on temporal event forecasting through LLMs.
\end{abstract}



\begin{keyword}
Temporal Event Forecasting \sep Temporal Knowledge Graph \sep Large Language Model \sep Retrieval Augmented Generation

\end{keyword}

\end{frontmatter}

Temporal event forecasting aims to predict future events based on observed historical events ~\citep{survey-event-forecasting}. This is a fascinating task that paves the way for humans to master the operating rules of the world, meanwhile, it also possesses significant practical and application value, such as offering early warning to critical events like civil unrest or regional conflicts ~\citep{GDELT,ICEWS,Glean}. Given the great values and impacts, temporal event forecasting has garnered increasing interest in both academic and research communities, and various studies have been conducted in recent years ~\citep{TKG-survey,LoGo}.
\begin{figure}
    \centering
    \includegraphics[width=\linewidth]{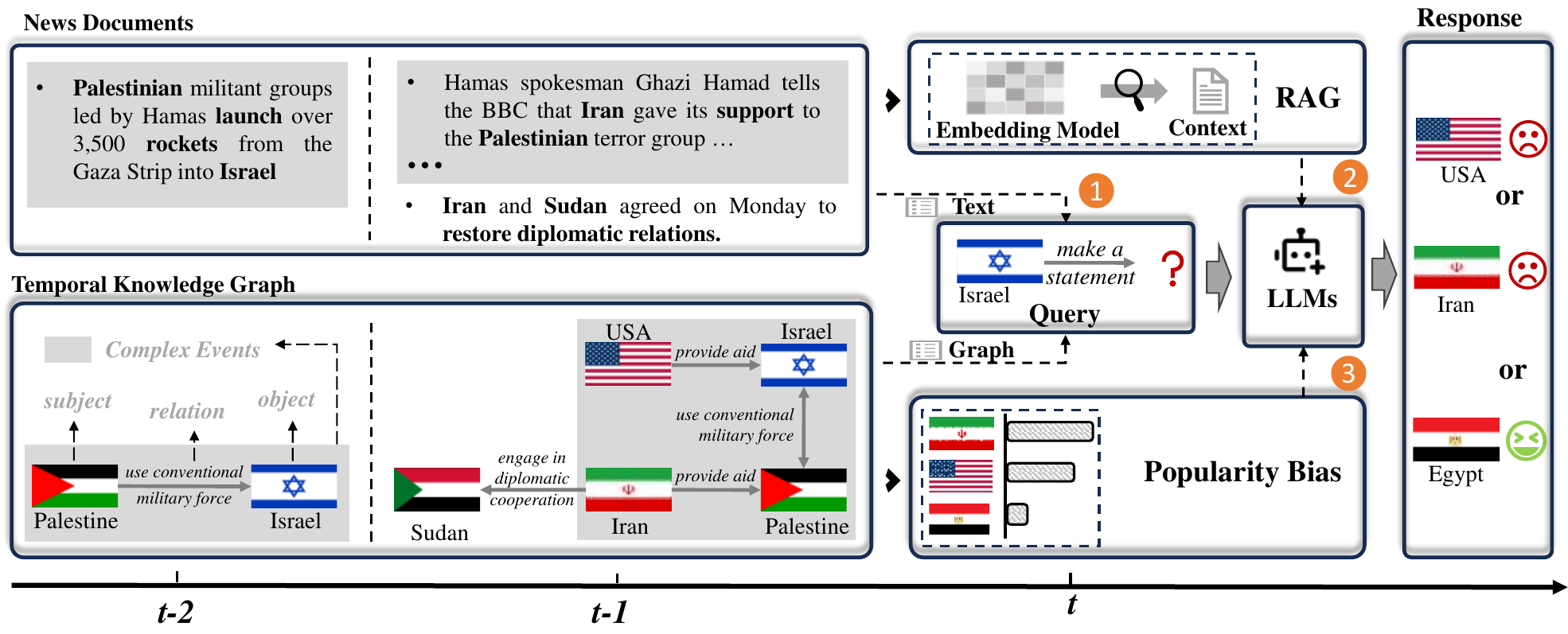}
    \caption{Illustration of leveraging LLM for temporal event forecasting. Given the complex event Israeli-Palestinian conflict , three formats of historical event representations, \ie text (top side), graph (bottom side), or graph-text (both), can be fed into the LLMs, and the LLMs are expected to answer certain input questions about what will happend in the future.}
    \label{fig: introduction}
\end{figure}
Among these studies, the representative formulations of temporal events can be broadly categorized into three formats~\citep{survey-event-forecasting}: graph-based, text-based, and graph-text hybrid. Specifically, the graph-based approach represents each event, \ie the so-called atomic event ~\citep{ConvE, ConvTransE,RGCN}, in a structured format, \ie a quadruple $(s, r, o, t)$, where $s$, $r$, $o$, and $t$ corresponds to the subject, relation~\footnote{Relation refers to the type of the atomic event.}, object, and timestamp, respectively. As illustrated in Figure \ref{fig: introduction}, each atomic event(\eg (Palestine, use conventional military force, Israel, t-2)) is extracted from a textual document(\eg Palestinian militant groups led by Hamas launch over 3,500 rockets from the Gaza Strip into Israel.
) , and multiple related atomic events form a complex event\citep{theFuture,SeCoGD,LoGo}, such as the Israeli-Palestinian conflict. Following such a structured representation, which is also termed as Temporal Knowledge Graph (TKG), various methods~\citep{RENET, REGCN, EvoKG, MA2024103848, LoGo} that target at modeling the temporal and relational patterns have been proposed and achieved remarkable progress~\citep{TKG-survey}. In contrast, text-based methods, such as event script prediction~\citep{scriptLearning} or forecast question answering (ForecastQA)~\citep{ForecastQA}, are characterized by directly consuming textual representations, which encapsulate more fine-grained details and contexts. Such detailed and contextual information is often ignored by graph-based methods since they are either not included in the ontology or not extracted by the information extraction system. Consequently, text-based event forecasting~\citep{ForecastQA,scriptLearning} concentrates more on the capabilities of natural language understanding and reasoning. The third branch, \ie graph-text hybrid method, aims to take advantage of the merits from both formats: structured reasoning and fine-grained information. Nevertheless, existing graph-text hybrid methods~\citep{Glean,CMF,SeCoGD} primarily treat text as side information to be integrated into the graph-based backbones, without truly conducting reasoning and forecasting on the texts themselves.

With the striking success of ChatGPT~\footnote{https://chat.openai.com}, LLMs have exhibited remarkable effectiveness across a wide range of tasks~\citep{GPT, RLHF, LLaMA} as well as event forecasting~\citep{PPT, GPT-NeoX-ICL, GENTKG, Chain-of-History}. These work pioneer in applying LLMs to the task of event forecasting through in-context-learning (ICL)~\citep{GPT-NeoX-ICL}, instruction tuning~\citep{PPT, Chain-of-History}, and retrieval augmented generation (RAG)~\citep{ToG, GENTKG} However, most of these approaches focus on discrete graph reasoning, thereby neglecting the broad contextual information inherent in natural language. Integrating the raw texts, from which the structured events are extracted, into the LLM-based forecasting processing is a natural and rational direction. 
Considering event forecasting is mostly related to critical scenarios such as international relationships and domestic stability, trustability and explainability are essentially required. Solely relying on the internal knowledge of LLMs would suffer from the problem of hallucination~\citep{GPT-NeoX-ICL, liang2024synergizing}. While RAG~\citep{RAG, sun-etal-2023-chatgpt} presents a promising solution, current graph-based RAG methods~\citep{GENTKG, ToG} are prone to noisy events obtained by unreliable event extraction systems. 
Therefore, the performance of text-based or graph-text hybrid RAG on temporal event forecasting is worth further exploration.

Furthermore, previous works~\citep{SeCoGD} have identified that event forecasting suffers from severe popularity bias. Traditional graph-based methods, which rely on Graph Neural Networks (GNNs) or Temporal Logical Rules (TLR) to capture relational and temporal patterns from historical events, tend to prioritize popular entities. This bias becomes particularly problematic when predicting entities that are either appearing for the first time or are infrequently encountered~\citep{GRLC}, such as Egypt in Figure \ref{fig: introduction}. In contrast, LLM-based methods have the potential to incorporate additional contextual information, potentially improving event forecasting for lesser-known entities. However, the extent to which LLMs, with their extensive world knowledge, are influenced by this bias remains uncertain.

To address this gap, we propose a systematic investigation into the capabilities LLMs in temporal event forecasting. Nevertheless, the main obstacle lies in the lack of well-established benchmark datasets. Most existing works are using graph-only datasets, \eg GDELT~\citep{GDELT} and ICEWS~\citep{ICEWS}, or text-only datasets, \eg ForecastQA~\citep{ForecastQA}. There are some text-enriched event graph datasets~\citep{SeCoGD,LoGo}, which enrich an existing structured TKG dataset with the raw texts that are downloaded through the news article URLs offered by GDELT. Unfortunately, the structured events in these datasets are highly noisy, because they are either from the original GDELT dataset~\citep{SeCoGD} or extracted using poor-performing systems~\citep{LoGo}, with only about 50\% accuracy as demonstrated in Section~\ref{subsec:dataset_construction}.

In this work, to conduct the evaluation, we first construct an exploratory benchmark data based on a previous dataset MidEast-TE~\citep{LoGo} and name our dataset as MidEast-TE-mini (short as MidEast-TE-m). Specifically, we sample a subset from the MidEast-TE~\citep{LoGo} and extract structured events using the state-of-the-art (SOTA) LLM GPT-4~\footnote{https://platform.openai.com/docs/models/gpt-4-and-gpt-4-turbo}. Based on this dataset, we design a series of baseline methods to evaluate the performance of LLMs on temporal event forecasting.
We are particularly interested in the following research questions: 
(1) \textbf{How do LLMs perform across different formulations of event forecasting?}
(2) \textbf{How does RAG perform on event forecasting? }
(3) \textbf{How do LLMs perform regarding the popularity bias in event forecasting?}
To answer these research questions, we conduct extensive experiments to verify the functionality of every designed component. The experimental results unveil multiple interesting phenomena, which not only deepen our understanding of LLM-based event forecasting methods but also surface several interesting research directions. The main contributions of this work are as follows:
\begin{itemize}
    \item To the best of our knowledge, we are among the first to systematically evaluate the performance of LLMs in text-involved temporal event forecasting. 
    \item To facilitate the evaluation, we construct a benchmark dataset with both raw texts and high-quality structured events. 
    \item We design a list of baseline methods and LLM-based methods catering to multiple evaluation criteria. Extensive experiments have unveiled meaningful insights and led to several noteworthy and valuable directions for future research. 
\end{itemize}

\section{Preliminary}
We first introduce the problem formulation of temporal event forecasting. Then, we present the dataset construction pipeline as well as the statistics and human evaluation results of our dataset.
\subsection{Problem Formulation}
Conventional temporal event forecasting methods define each event as a quadruple $(s, r, o, t)$, referred to as an atomic event. Following recent studies~\citep{theFuture, LoGo}, atomic events are grouped into complex events (CE), which include a set of correlated atomic events and represent major events that span over longer time spans and cover more entities. For example, all the events within the red dashed box in Figure \ref{fig: introduction} illustrate a complex event, specifically the Israeli-Palestinian conflict. Consequently, we define an atomic event at timestamp $t$ as $(s, r, o, t, c)$, where $s\in\mathcal{E}$, $r\in\mathcal{R}$, and $o\in\mathcal{E}$ represent the subject, relation, and object, respectively; $\mathcal{E}$ and $\mathcal{R}$ denote the entity and relation sets, and $c$ represents the complex event type. Notably, the timestamp $t$ is converted into a relative timestamp. All atomic events at the same timestamp $t$ form an knowledge graph, denoted as $G_t={(s_n, r_n, o_n, t, c_n)}_{n=1}^{N}$, where $(s_n, r_n, o_n, t, c_n)$ represents the $n$-th event, and $N$ is the total number of events.

In addition to the traditional graph-based representation, we aim to incorporate raw news text into temporal event forecasting. Specifically, each knowledge graph $G_t$ is extracted from a list of news documents $D_t=\left [ d_1, d_2, \ldots, d_k \right ]_{k=1}^K \in \mathcal{D}$, where $d_k$ denotes the $k$-th document at timestamp $t$ and $\mathcal{D}$ denotes the set of documents. It is important to note that the same document $d$ can contribute to multiple events. Given the historical knowledge graphs or news documents prior to $t$, denoted as $\mathbf{G}_{<t} = \left \{ G_0, G_1, \ldots, G_{t-1} \right\}$ and $\mathbf{D}_{<t} = \left \{ D_0, D_1, \ldots, D_{t-1} \right\}$, respectively, and a query, represented as $(s,r/o,t)$, temporal event forecasting aims to predict the missing object or relation.

\subsection{Dataset Construction}~\label{subsec:dataset_construction}
We first describe the data source that we utilized for curating the dataset. Then we present the dataset construction pipeline, followed by the dataset construction. 

\subsubsection{Data Source}
We construct our dataset based on a subset of MidEast-TE\citep{LoGo}, therefore, we name our dataset as MidEast-TE-mini, short as MidEast-TE-m. The original MidEast-TE dataset includes both structured atomic events and news articles, which have already been clustered into complex events. Given the large scale of MidEast-TE, we sample a subset of complex events from it and keep the associated news articles as the original documents to build our dataset. 
Specifically, the time spans of the complex events in MidEast-TE are of high divergence, ranging from several days to more than three months. To reduce the potential bias introduced by outliers, we just sample 120 complex events whose time spans are within 40-60 days. For the shorter and longer complex events, we do not cover them in this exploratory dataset and leave them for future work.
\begin{figure}
    \centering
    \includegraphics[width=1\linewidth]{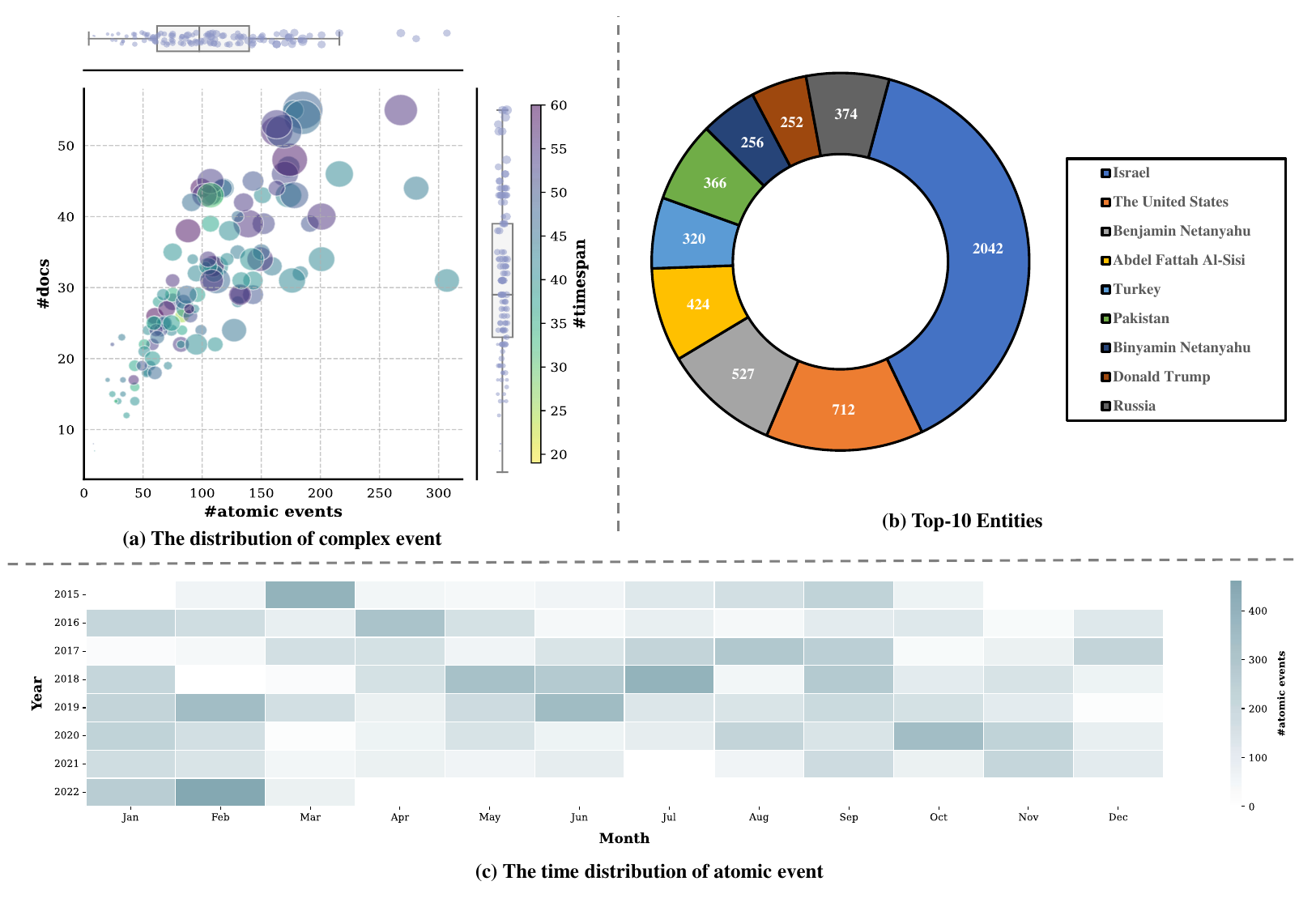}
    \caption{The data distribution of MidEast-TE-m.}
    \label{fig: distribution}
\end{figure}
\subsubsection{Construction Pipeline}
The overall pipeline of the dataset construction involves two consecutive components: Event Extraction and Entity Linking.

\noindent \textbf{Event Extraction.} We instruct GPT-4 to perform the task of event extraction. Since the set of event types is quite large ($>200$), putting all the event definitions into prompts yields high cost and may decrease the performance. Therefore, we follow the previous approach~\citep{LoGo} and conduct a multi-level event extraction given the three-layer hierarchical structure of CAMEO~\citep{CAMEO} ontology. We include the description of each event type defined in CAMEO in the prompt. During the first-level extraction, we break down each article into multiple sentence-level trunks, each with around 150 tokens. Then, we extract the first-layer atomic events from each trunk. 
After recognizing atomic events based on first-level event types, for the second-level extraction, we group 15 atomic events with the same first-level event type and their respective trunks into a single prompt. This approach aims to maximally share the prompt while maintaining the high-quality extraction. 
The same procedure is then applied to the third-level extraction. 

\noindent \textbf{Entity Linking.} Due to the absence of a predefined entity set in our event extraction process, resulting in repeated while diverse-formatted entities. 
Subsequently, we employ an entity linking step using GPT-4. 
Initially, we apply K-means to cluster all original entities into multiple groups. Then, we batch the entities from each cluster and ask GPT-4 for entity linking. 

\subsubsection{Dataset Statistics}
We split the dataset temporally and subsequently present its distribution. We then assess the quality of the benchmark dataset.

\noindent \textbf{Data splits.}
\begin{table}[t]
\begin{center}
\caption{Statistics of our curated dataset MidEast-TE-m.}
\label{tab:dataset_3}
\vspace{3mm}
\resizebox{0.6\textwidth}{!}{
    \begin{tabular}{lc cccc}
        \toprule
        \textbf{Dataset} & \textbf{\#atomic events} & \textbf{\#CEs} & $|\mathcal{E}|$ & $|\mathcal{R}|$ & \textbf{\#docs}\\
        \midrule
        \textbf{train} & 8,999 & 88 & 4,529 & 254 & 2,647\\
        \textbf{val} & 1,777 & 19 & 989 & 191 & 473\\
        \textbf{test} & 1,766 & 18 & 1,146 & 198 & 572\\
        \midrule
        \textbf{total} & 12,542 & 120 & 5,909 & 267 & 3,692\\
        \bottomrule
    \end{tabular}
}
\end{center}
\end{table}
Table~\ref{tab:dataset_3} details the statistics of our curated dataset MidEast-TE-m, where \#atomic events, \#CEs, and \#docs represent the number of atomic events, complex events, and news documents, respectively. Additionally, $|\mathcal{E}|$ and $|\mathcal{R}|$ refer to the number of entities and relations. We split the dataset into train, validation, and test sets in a temporal manner. Specifically, we use the atomic events in the last year for testing, the second-to-last year for validation, and the rest about five years for training. Note that the number of atomic events is not the same as the number of documents, as a single news document may correspond to multiple events.

\noindent \textbf{Data Distribution.} Figure \ref{fig: distribution} (a) presents the distribution of the complex events, where each point represents a specific complex event. The x-axis indicates the number of atomic events contained within each complex event, while the y-axis represents the number of associated news documents. The size of each scatter point is proportional to the number of event entities within the complex event, and the color of the points corresponds to the time span of the events, with lighter colors representing shorter time spans and darker colors indicating longer durations. Apparently, more complex events typically encompass a larger number of atomic events and associated news texts. The variation in point size and color further highlights the diversity of complex events in terms of time span and the number of entities involved. In addition, the box plots above and to the right provide insights into the distribution of atomic events and news texts, respectively, within complex events.
The Top-10 entities with the highest frequency in the dataset are shown in Figure \ref{fig: distribution} (b). 
In order to display the temporal trends of our dataset, Figure \ref{fig: distribution} (c) illustrates the temporal distribution of atomic events, news documents, and complex events across different months (x-axis) and years (y-axis). The color intensity reflects the number of events occurring in each time period. We can see that the temporal distribution of MIDEAST-TE-m is relatively uniform.

\begin{table}[!htbp]
    \caption{Error analysis of the event extraction results in different datasets.}
    \vspace{3mm}  
    \centerline{
    \resizebox{0.7\textwidth}{!}{\begin{tabular}{lcccccc}
        \toprule
        \multirow{2.4}{*}{\textbf{Dataset}} & \multirow{2.4}{*}{\shortstack{\textbf{\#atomic} \\ \textbf{events}}} & \multirow{2.4}{*}{\textbf{Acc.(\%)}} & \multicolumn{3}{c}{\textbf{error type (\%)}} \\
        \cmidrule{4-6}
        & & & \textbf{time} & \textbf{relation} & \textbf{entity} \\
        \midrule
        \textbf{GDELT-TE} & 148 & 29.73 & 3.85 & 31.73 & 64.42 \\
        \textbf{MidEast-TE} & 35 & 55.56 & 0 & 92.86 & 7.14 \\
        \midrule
        \textbf{MidEast-TE-m(ours)} & 78 & 72.60 & 16.67 & 27.78 & 55.56 \\
        \bottomrule
    \end{tabular}}}
    \label{tab:dataset_eval}
\end{table}

\noindent \textbf{Human Evaluation.} In order to evaluate the quality of the benchmark dataset, we conduct human evaluation. 
Specifically, given a news article and the extracted events, we ask the human evaluator to tell whether the extracted events are valid or not based on the original article.

We randomly sample 20 documents from the document set of the dataset, and conduct human evaluation based on the following criteria:
\begin{itemize}
    \item Time: the atomic events extracted are events that have already occurred or are currently happening, rather than future events.
    \item Relation: the extracted atomic events appear in the original text and faithfully reflect the semantics of the original text.
    \item Entity: the extracted entities are concrete and real-world entities, and they appear in the original news article.
\end{itemize}
To be noted, for every extracted atomic event, we evaluate it once by going through the three criteria sequentially. For example, if we identify the \textit{Time} is incorrect, we stop the checking of the following two criteria.  Table~\ref{tab:dataset_eval} shows the accuracy and error types of the event extraction results, including both our dataset MidEast-TE-m and previous datasets of MidEast-TE~\citep{LoGo} and GDELT-TE~\citep{GDELT}. To be noted, MidEast-TE is constructed using Vicuna-7b~\citep{vicuna2023}, while GDELT-TE is curated using proprietary extraction system.
Clearly, the dataset constructed by our pipeline is the most accurate one. In our dataset constructed using GPT-4, there are the fewest extraction errors,  which makes the following forecasting research more valid and reliable. However, we admit that an overall accuracy of 72.6\% is still unsatisfactory in practice, and further efforts will be devoted to improved event extraction performance.

\section{Methods}

\begin{figure*}
    \centering
    \includegraphics[width=0.99\textwidth]{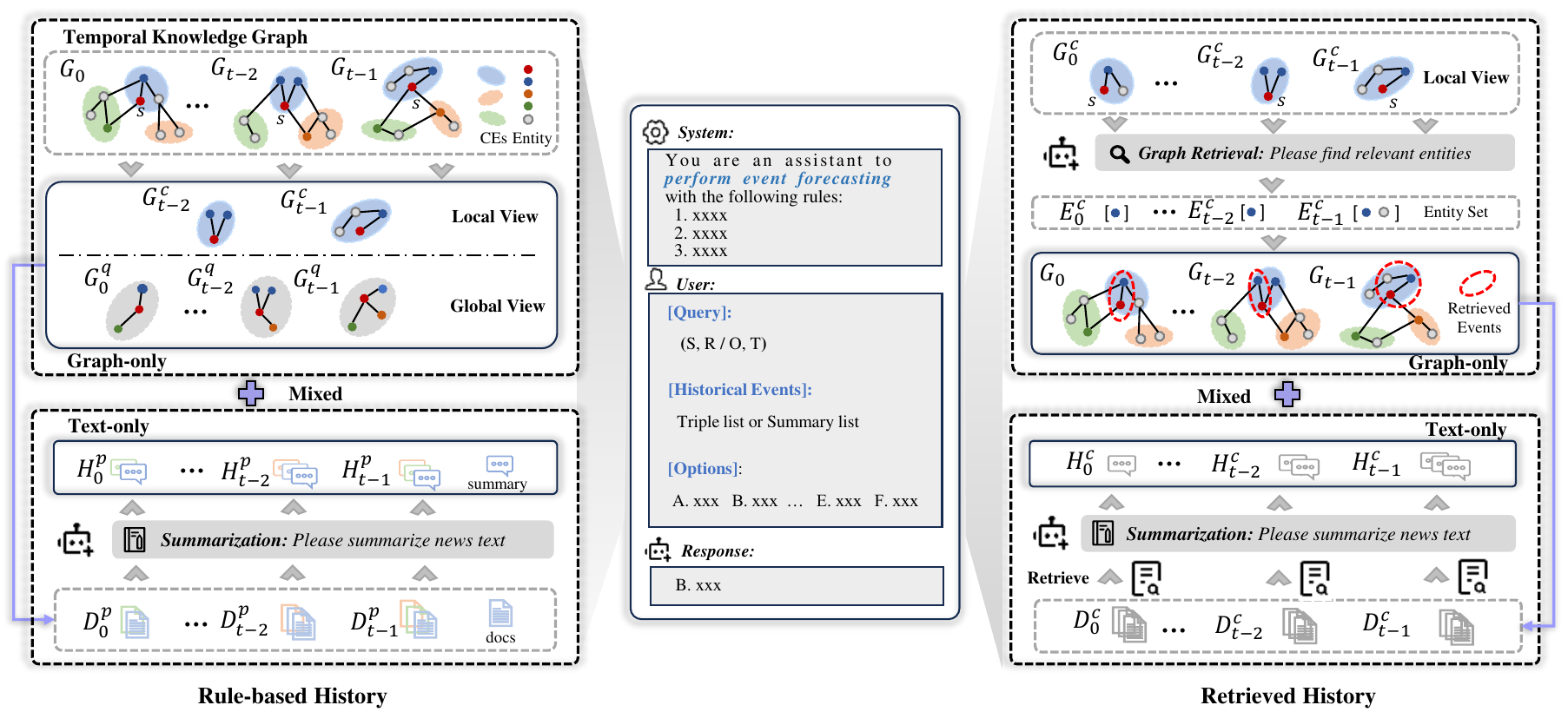}
    \caption{
   Illustration of rule-based history and retrieved history. The rule-based history is constructed by a set of predefined rules. In contrast, the retrieved history dynamically searches context from the temporal knowledge graph or news documents according to the current query.}
    \label{fig:framework}
\end{figure*}

We systematically designed a series of LLM-based methods for event forecasting, categorized into three streams: graph-only, text-only, and graph-text (mixed) methods. Notably, we focus on LLM-based methods, while non-LLM baselines are described in Section~\ref{subsec:exp_setting}. Additionally, we developed two approaches for constructing event history to assist LLM reasoning in temporal event forecasting, as illustrated in Figure~\ref{fig:framework}. These approaches include rule-based history and retrieved history.
\subsection{Graph-only Methods}

Graph-only methods utilize discrete temporal knowledge graphs as inputs, leveraging the inherent domain knowledge of LLMs to reason over historical events and make forecasts. To enhance contextual understanding, we developed two methods for constructing event history relevant to the query event. 

    \noindent \textbf{Rule-based History.}
    Inspired by the previous work~\citep{LoGo}, we construct event history from local and global views. In the global view, we reconstruct the knowledge graph $\mathbf{G}_{<t}^q = \left \{ G_0^q, G_1^q, \ldots, G_{t-1}^q \right\}$  based on the connections of the query subject $s$ within the original temporal knowledge graph (TKG), where $G_{t'}^q = \left \{ (s, r_{i}, o_{j}, c_{k}, t')| (s, r_{i}, o_{j}, c_{k}, t') \in G_{t'} \right\}$ represents historical events at timestamp $t'$ with the same subject as the query. In the local view, we extract a subgraph to provide fine-grained contextual information for LLM reasoning. Formally, complex knowledge graph $\mathbf{G}_{<t}^{c}= \left \{ G_0^c, G_1^c, \ldots, G_{t-1}^c \right\}$, where $G_{t'}^c= \left \{ (s_{i}, r_{j}, o_{k}, c, t')| (s_{i}, r_{j}, o_{k}, c, t') \in G_{t'} \right\}$ represents historical events at timestamp $t'$ in the same complex event $c$. Due to length constraints on event history, we retain only the most recent Top-K events for $\mathbf{G}_{<t}^{q}$ and events from the past two days for $\mathbf{G}_{<t}^{c}$ to construct the event history.
    
    \noindent \textbf{Retrieved History. }
    Unlike rule-based history, retrieved history typically leverages the powerful capabilities of LLMs to identify relevant contextual information. Firstly, we selected the historical events from the past days within $\mathbf{G}_{<t}^{c}$ to form an entity candidate set, which represents the potential actors of the query event. This candidate set is then input to the LLM to filter out irrelevant entities. The instruction of graph retrieval is detailed in Table ~\ref{tab: prompts of graph retrieval}. Once the relevant entities, denoted as $\mathbf{E}_{<t}^{c}= \left \{ E_0, E_1, \ldots, E_{t-1} \right\}$, are identified, we extract the historical events that containing $e^{c} \in \mathbf{E}_{<t}^{c}$ to construct the event history.

\begin{table*}[!htbp]
\centering

\caption{ Instruction of graph retrieval.}
\vspace{3mm}
\label{tab: prompts of graph retrieval}
\begin{tabularx}{\linewidth}{p{\linewidth}}
\toprule
You are an assistant to find relevant entities with the following rules:

\textbf{1.} [Subject] represents the event subject in a specific event. [Candidate Set] represents a list of entities.

\textbf{2.} You need to select the entities that may be relevant to [Subject].
\\
\bottomrule

\end{tabularx}
\end{table*}
\subsection{Text-only Methods}
Text-only methods utilize raw textual data for forecasting. 
Each atomic event originates from a specific news document. Similar to graph-only methods, relevant event history is obtained through rule or retrieval.

\noindent \textbf{Rule-based History.}
Generally, news documents provide more detailed contextual information for temporal event forecasting, making the exploration of textual information related to the current event a critical step. Following the graph-only method, we identify the historical events from $\mathbf{G}_{<t}^{q}$ and $\mathbf{G}_{<t}^{c}$ and construct the news documents set $\mathbf{D}_{<t}^p =  \left \{ D_0^p, D_1^p, \ldots, D_{t-1}^p \right\}$, from which these historical events can be extracted. Given that news documents often contain a substantial amount of irrelevant information regarding the query event, we employ LLMs to summarize these documents, producing concise summaries $\mathbf{H}_{<t}^p =  \left \{ H_0^p, H_1^p, \ldots, H_{t-1}^p \right\}$ that retain only the core thematic information. 
Compared to methods that rely solely on temporal knowledge graphs, the text-only method focuses on predicting the missing objects based on historical news summaries.

\noindent \textbf{Retrieved History. }
Inspired by the recent studies of RAG~\citep{ToG,GENTKG}, we construct the news document set $\mathbf{D}_{<t}^c =  \left \{ D_{t-h}^c, \ldots, D_{t-2}^c, D_{t-1}^c \right\}$ from the past $h$ days within the same complex event as the query event. To obtain the fine-grained textual information, these news documents are divided into smaller chunks, which are then encoded into vector representations using an embedding model. The query subject is utilized to compute the similarity scores and find relevant news text. The retrieved news texts are subsequently sorted by time and similarity scores, with historical texts that are excessively outdated relative to the event timestamp being filtered out. Similarly to rule-based history, we employ LLM to generate concise summaries $\mathbf{H}_{<t}^c =  \left \{ H_0^c, H_1^c, \ldots, H_{t-1}^c \right\}$ for the filtered news text set, serving as contextual information to predict missing object in the current event.

\subsection{Graph-and-Text (Mixed) Methods}
Graph-and-text methods leverage both graph and text data as input. 
Differing from the above two methods, the graph-and-text method combines TKG and news documents, with the TKG supplying structured information and the news documents providing fine-grained background information. The two specific implementations of this method are as follows:

\noindent \textbf{Rule-based History.}
The rule-based history with graph-and-text integrates graph-only history with text-only history to form a comprehensive graph-and-text history. The input to the LLM includes not only historical events but also their corresponding news summaries. The graph history provides pertinent details about the temporal knowledge graph's structure, while the text history offers detailed background information in a more granular manner.

\noindent \textbf{Retrieved History. }
Rather than directly concatenating graph-only history with text-only history, the mixed retrieval method retrieves news text not only based on the subject of the query event but also the relevant entity set $\mathbf{E}_{<t}^{c}$, thereby enriching the text exploration with structural graph information. The retrieved news text, in turn, better assists LLM in conducting structured reasoning. To address the limitation of context length, we also transform the retrieved news text into summaries by LLMs.
\begin{table*}[!htbp]
\centering
\small
\caption{Instruction of reasoning.}
\vspace{3mm}
\label{tab: prompts of event forecasting}
\begin{tabularx}{\linewidth}{p{\linewidth}}
\toprule
 
 You are an assistant to perform event forecasting with the following rules:

\textbf{1.} The atomic event is the basic unit describing a specific event, typically presented in the form of a quadruple (S, R, O, T), where S represents the subject, R represents the relation, O represents the object, and T represents the relative time.

\textbf{2.} Complex Event, which is composed of a set of atomic events, describes the temporal evolution process of multiple atomic events.

\textbf{3.} Please remember the meanings of the following identifiers: 

\textbf{[Query]} represents the event to be predicted in the form of (S, O, T). 

\textcolor{red}{\textbf{[Nearest Events]}}represents a list of (\textit{summaries} / \textit{atomic events} / \textit{summaries and atomic events}) in the complex event that is relatively closer in time to the predicted event. 

\textcolor{red}{\textbf{[Further Events]}} represents a list of (\textit{summaries} / \textit{atomic events} / \textit{summaries and atomic events}) in the complex event that is relatively further in time from the predicted event. 

\textcolor{red}{\textbf{[Related Events]}} encompasses the past (\textit{summaries}  / \textit{atomic events} / \textit{summaries and atomic events})  that contain the same subject or object as the question. 

\textcolor{blue}{\textbf{[Relevant Event]}} represents a list of (\textit{atomic events} / \textit{summaries and atomic events}) relevant to the query. 

\textcolor{blue}{\textbf{[Relevant News Text]}} represents background information about subject. 

\textbf{[Options]} represents the candidate set of the missing object.

\textbf{4.} Given a query of (S, O, T) in the future and the list of historical events until t, event forecasting aims to predict the missing object.
\\
\bottomrule

\end{tabularx}
\end{table*}

\subsection{Prompt Design}

To effectively guide LLMs in performing temporal event forecasting, we design a specific prompt for this task, which consists of an instruction and an input component. As shown in Table~\ref{tab: prompts of event forecasting}, the instruction includes the definitions of atomic and complex events, the meanings of the identifiers used in the input, and an overview of the temporal event forecasting task. Notably, red identifiers represent those used for the rule-based history, while blue identifiers denote those used for the retrieved history. As depicted in Figure~\ref{fig:framework}
, the input comprises the query, historical events, and the options. The format for the input of historical events is detailed in Table~\ref{tab: prompts of history}.

\begin{table*}[!htbp]
\centering
\small
\caption{Input format of historical events.}
\vspace{3mm}
\renewcommand{\arraystretch}{1.5}
\label{tab: prompts of history}
\begin{tabular}{l p{0.7\linewidth}}
\toprule
\textbf{Input Format} & \textbf{Historical Events} \\
\midrule
\textbf{Graph-only}
 &
 \multicolumn{1}{m{0.7\linewidth}}{
(Pakistan Foreign Ministry, Cooperate economically, Egyptian counterpart officials, 2190); 
}
\\
\hline
\textbf{Text-only}
 &
 \multicolumn{1}{m{0.7\linewidth}}{
[Date]2190:

Pakistani Foreign Minister Shah Mahmood Qureshi expressed great compatibility...
}

\\
\hline

\textbf{Mixed}
 &
 \multicolumn{1}{m{0.7\linewidth}}{
[Date]2189:

Foreign Minister Shah Mahmood Qureshi visited Egypt...

(Shah Mahmood Qureshi, Make optimistic comment, Pakistan); 
}
\\

\bottomrule
\end{tabular}
\end{table*}
\subsection{Fine-tuning}
Previous work has demonstrated the effectiveness of fine-tuning LLMs on temporal event forecasting\citep{GENTKG,Chain-of-History}. In order to comprehensively evaluate the impact of fine-tuning, we adopt the Low-Rank Adaptation (LoRA)\citep{QLoRA} for open-source LLMs. The core idea of LoRA is to decompose the weight matrices $\mathbf{W}$ of a pre-trained model into a sum of a low-rank matrix and a residual matrix, typically expressed as $\mathbf{W}'=\mathbf{W}+\bigtriangleup \mathbf{W}$, where where 
$\mathbf{W}'$ is the updated weight matrix after fine-tuning, and $\bigtriangleup \mathbf{W} = \mathbf{A}\mathbf{B}$ is the low-rank adaptation matrix; $\mathbf{A}$ and $\mathbf{B}$ are two smaller matrices that allow the model to adapt to new tasks. Due to limitations in training data and time constraints, we restricted fine-tuning to the rule-based history. Specifically, we integrated the rule-based history into the designed prompt and computed the cross-entropy loss between the predicted object and the actual answers for optimization.

\section{Experiments}
Implementing these methods, we are particularly interested in answering the following research questions: 
\begin{itemize}
    \item \textbf{RQ1: } What is the overall performance of LLM-based methods, with various input modalities and forecasting objectives?
    \item \textbf{RQ2: } Is RAG helpful and how do various retrieval-relevant settings affect the performance?
    \item \textbf{RQ3: } How does the popularity bias (long-tail) issue affect the forecasting performance?
\end{itemize}

\subsection{Experimental Settings}~\label{subsec:exp_setting}
To evaluate various methods, we conduct experiments on our proposed dataset MidEast-TE-m, as described in Section~\ref{subsec:dataset_construction}. Following previous studies, we adopt \textit{Accuracy} (Acc) as the evaluation metric. For non-zero shot methods, we train each model on the training set, select the best-performing model based on the validation set, and then evaluate it on the test set. In contrast, for zero-shot approaches, we directly evaluate on the test set without any additional training.
\subsubsection{Compared Methods}
In addition to LLM-based methods, we implement a list of representative graph-based methods, described as below:
\begin{itemize}
    \item \textbf{ConvTransE~\citep{ConvTransE}:} it is a static knowledge graph (KG) representation learning method, which uses both a convolutional neural network and a translational operation to capture the patterns within triplets. 
    \item \textbf{RGCN~\citep{RGCN}:} this is also a static KG representation learning model, which leverages graph convolutional neural network to model various relations among entities. 
    \item \textbf{RE-GCN~\citep{REGCN}:} REGCN is a SOTA method for TKG, which leverages GNN and RNN to capture both relational and temporal patterns, respectively.
    \item \textbf{CMF~\citep{CMF}:} CMF proposes a context-based feature fusion method, which integrate multilevel features including event frequency, news documents and event graphs.
    \item \textbf{LoGo~\citep{LoGo}:} LoGo is the SOTA method for the temporal complex event (TCE), which employs two RT-GNN modules to model both local (within complex event) and global (across all complex events) contexts. To be noted, only LoGo leverage the complex event information, while the above three methods do not use it.
    \item \textbf{ForecastQA~\citep{ForecastQA}:} ForecastQA simulate the the forecasting scenario on temporal news documents and design a method based on pretrain language models to make a forecasting judgement with the retrieved documents.
    \item \textbf{CoH~\citep{Chain-of-History}:} It is a LLM-based method that construct the event history by designed rules and employ fine-tuning to understand textual graph information.
    \item \textbf{GenTKG~\citep{GENTKG}:} this is a retrieval-augmented generation framework that incorporates temporal logical rules into the retrieval process. Additionally, fine-tuning is employed to adapt LLM for the task of temporal event forecasting.
    
\end{itemize}

\begin{table*}
\centering
\caption{Performance (Accuracy) comparison of LLM-based methods and non-LLM methods. "N/A" stands for "Not Applicable".}
\label{tab:overall_performance}
\renewcommand{\arraystretch}{1.3}
\resizebox{\linewidth}{!}{
\begin{tabular}{lllccc  ccc }
\toprule
\multirow{2}{*}{\textbf{Model Type}} & \multirow{2}{*}{\textbf{Model}} & 
\multirow{2}{*}{\textbf{Backbone}} & \multicolumn{3}{c}{Object Entity Prediction} & \multicolumn{3}{c}{Relation Prediction} \\
\cmidrule{4-9}
 & & & \textbf{Graph-only} & \textbf{Text-only} & \textbf{Mixed} & \textbf{Graph-only} & \textbf{Text-only} & \textbf{Mixed} \\ 

\midrule
\multirow{6.4}{*}{Non-LLM} & \textbf{ConvTransE~\citep{ConvTransE}} & \multirow{5.4}{*}{GNN}
 &0.3737 &N/A &N/A &0.7327 &N/A &N/A \\
& \textbf{RGCN~\citep{RGCN}} 
&   
&0.3777 &N/A &N/A 
&0.7203 &N/A &N/A \\
& \textbf{RE-GCN~\citep{REGCN}} 
&  
&0.3879 &N/A &N/A 
&0.7333 &N/A &N/A  \\
& \textbf{CMF~\citep{CMF}} 
& 
&N/A &N/A &\underline{0.3783} 
&N/A &N/A &\underline{0.7265}  \\
& \textbf{LoGo~\citep{LoGo}} 
& 
&0.3969 &N/A &N/A 
&0.7406 &N/A &N/A \\
& \textbf{ForecastQA~\citep{ForecastQA}} 
& Bert
&N/A &\underline{0.3901} &N/A 
&N/A &\underline{0.7389} &N/A \\

\midrule
\multirow{2}{*}{Zero-shot} 
& \textbf{rule-based history} 
& \multirow{2}{*}{gpt-3.5-turbo}  
&0.3290 &0.3233 &0.3154 
&0.5515 &0.4807 &0.5464 \\
& \textbf{retrieved history} 
& 
&0.3533 &0.3228 &0.3539 
&0.5498 &0.4637 &0.5402 \\

\midrule
\multirow{3}{*}{Fine-tune}
& \textbf{CoH~\citep{Chain-of-History}}
& \multirow{2}{*}{Llama2-7b}  
&\underline{0.4856} &N/A &N/A 
&0.7763 &N/A &N/A \\
& \textbf{GenTKG~\citep{GENTKG}} 
&  &0.4785 &N/A &N/A 
&\underline{0.7907} &N/A &N/A \\ \cmidrule{2-9}
& \textbf{rule-based history} 
& \multirow{1}{*}{Vicuna-7b} 
&\textbf{0.5271} &\textbf{0.5266} &\textbf{0.5798} &\textbf{0.8023} &\textbf{0.8103} &\textbf{0.8058} \\

\bottomrule
\end{tabular}}
\end{table*}

\subsubsection{Implementation Details}
Owing to the large entity set and limited context length of LLMs, we simplify the event forecasting settings as an Multi-Choice Questions(MCQ) problem and adopt negative sampling strategies to construct candidate sets. Specifically, we randomly sample two entities from $G_{t-1}^c$, two entities from $G_{t-2}^c$, and one entity from $\mathbf{G}^q$. As a result, the length of the object candidate set is set to 6. 
For the LLM-based method in this paper, we employ a variety of LLMs which include Llama2-7b~\footnote{https://huggingface.co/meta-llama/Llama-2-7b-chat}, Vicuna-7b~\footnote{https://huggingface.co/lmsys/vicuna-7b-v1.5}, and gpt-3.5-turbo~\footnote{https://platform.openai.com/docs/models/gpt-3-5-turbo}, where the temperature to 0 and the seed parameter to a fixed integer for reproducibility in our experimental setup.  The maximum token length of output is set to 256 to prevent invalid responses.

For the rule-based method, we generate corresponding summaries for all documents in advance with Vicuna-7b. Notably, we further apply fine-tuning (instruction tuning with QLoRA~\citep{QLoRA}) to these open-source LLMs. Since the retrieved history directly retrieves news text, we utilize gpt-3.5-turbo to generate summaries for the retrieved news text set and keep the retrieval models fixed without tuning.  The retrieval models include three different retrievers:  BM25~\citep{BM25}, Contriever~\citep{Contriever}, and LlamaIndex~\citep{LlamaIndex}.To ensure the fairness of the experiment, for all the baselines and LLM-based methods, we search historical lengths from $\{3, 7, 15, 30, 90\}$.
Moreover, considering the token length constraints of LLMs, we we limit the number of historical events and texts to 20 and 5, respectively.

\subsection{Performance Comparison (RQ1)}
We aim to compare the performance between various non-LLM methods and LLM-based methods, in terms of different input and output settings, as well as various negative sampling strategies.
\subsubsection{Performance \wrt Various Inputs and Outputs.} 
Table~\ref{tab:overall_performance} presents a comparison of overall performance for both object entity prediction and relation prediction tasks.From these results, we make the following observations. First, the LLM-based method proposed in this paper achieves the highest performance under the fine-tuned setting. These preliminary findings suggest that fine-tuning LLMs holds substantial promise for event forecasting, indicating that more sophisticated fine-tuning strategies warrant further investigation. Compared to the other fine-tuned LLM-based methods such as CoH and GenTKG, the introduction of complex events and news text provides more fine-grained information for temporal event forecasting. Moreover, enhanced with the retrieval modules, LLM-based methods in zero-shot setting can achieve not bad performance for both tasks of tail entity and relation prediction. Nevertheless, they still under-perform conventional non-LLM-based methods on all three formats of inputs, \aka Graph-only, Text-only, and Mixed, demonstrating that non-LLM-based methods remain competitive. Second, surprisingly, incorporating raw text into the input does not demonstrate positive effects in zero-shot setting. The performance on the text-only setting is obviously worse than the graph-only and mixed settings. This may be because the raw textual inputs are longer than the graph-only inputs, resulting in increasing difficulty in reasoning. Third, comparing the two forecasting tasks, the absolute performance of relation prediction is higher than object entity prediction by an obvious margin, indicating that forecasting entities are more challenging than relations in the domain of international cooperation and conflict events. The possible reasons are from two aspects. First, the entity set is much larger than the relation set, therefore, predicting entity is more difficult than relation given a larger candidate set. Second, when predicting relations, the input question includes both the subject and object entity, while the subject entity and relation are the input when predicting object entity. We deem that entities retain more information than relation in event forecasting, thus the question of relation prediction could offer more information and is easier. 

\subsubsection{Performance \wrt Backbone LLMs}
The capability of the backbone LLM is critical for robust performance across various tasks. In addition to the commercial proprietary LLMs, we are also interested in the performance of open-source LLMs. These open-source LLMs can be adapted, re-configured, and more importantly, safer than commercial LLMs due to privacy issues. Accordingly, we select two widely adopted open-source LLMs, \ie Llama2-7b and Vicuna-7b, for evaluation. The results, summarized in Table~\ref{tab:effects_backbone}, reveal that both Llama2-7b and Vicuna-7b exhibit significantly lower performance than GPT-3.5-turbo under the zero-shot setting, highlighting the inherent capacity gap between smaller open-source models and more powerful commercial models. Nevertheless, Under the fine-tuning setting, all the open-source models exhibit significant performance gains. They are not only better than the zero-shot version counterparts but also outperform gpt-3.5-turbo by a large margin. This phenomenon indicates that supervised instruction fine-tuning is crucial for enhancing the ability of LLM to learn temporal relational patterns.
Additionally, the performance of Vicuna-7b outperforms Llama2-7b, implying that advanced LLM capabilities are more effective at capturing evolutionary patterns in historical events.
\begin{table}
\centering
\caption{Performance comparison \wrt different backbone LLMs and fine-tuning.}
\vspace{3mm}
\label{tab:effects_backbone}
\resizebox{0.8\linewidth}{!}{\begin{tabular}{llccc}
\toprule
\shortstack{\textbf{Model Type} \\ \textbf{(rule-based history)}} & \textbf{Backbone} & \textbf{Graph-only}  & \textbf{Text-only} & \textbf{Mixed} \\
\midrule
\multirow{3.4}{*}{Zero-shot}  
& \textbf{Llama2-7b}  & 0.2225 & 0.2639  &0.2157  \\
& \textbf{Vicuna-7b}  & 0.2027 &0.2384 & 0.2524 \\
& \textbf{gpt-3.5-turbo} & \textbf{0.3290} &\textbf{ 0.3233} & \textbf{0.3154}\\
\midrule
\multirow{2}{*}{Fine-tune}  
& \textbf{Llama2-7b}   & 0.5057 &0.4892 & 0.4977 \\
& \textbf{Vicuna-7b}  & \textbf{0.5271} & \textbf{0.5266} & \textbf{0.5798} \\
\bottomrule
\end{tabular}}
\end{table}

\subsubsection{Performance \wrt Various Negative Sampling Strategies} 
Given the large entity set and limited context length of LLMs, we simplify the event forecasting settings as an MCQ problem, \ie choose the correct answer from six options. However, previous settings, such as TKG works~\citep{REGCN,LoGo}, treat the whole entity (relation) set as options during forecasting, aligning well with the real application scenario of the open world. Consequently, our simplified setting of MCQ could inevitably induce bias \wrt different sampling strategies for negative (wrong) options generation. To investigate the impact of negative sampling strategy, we introduce two more strategies in addition to the default strategy \textit{history}, \ie 1) \textit{global}, which randomly sample entity from the entire entity set; and 2) \textit{generated}, which are newly generated by gpt-3.5-turbo and not same with the ground-truth answer~\footnote{The generated entities include some entities that are not in the current entity set, therefore, non-LLM methods are not applicable to this setting due to they cannot deal with new entities). Nevertheless, we do not force the generated entities to be totally new, which means the generated entities may be overlapped with existing entities.}. The results are shown in the right hand side of Table~\ref{tab:overall_performance}, from which we can observe several interesting findings. First, among all the three strategies, all the methods perform worst on the \textit{history} setting, showing that our default setting of sampling negative options from event history is the most challenging setting. Second, \textit{global} demonstrates the highest performance, showing to be the easiest setting. Meanwhile, \textit{generated} is in the middle, with a balanced level of difficulty. Finally and more interestingly, LLM-based models' performance on \textit{global} outperforms all the non-LLM methods. This may be because LLMs are superior in answering general easy questions with common knowledge, while less capable in dealing with hard questions in specific domains.

\begin{table}
\centering
\caption{Performance comparison \wrt various negative sampling strategies. "N/A" means "Not Applicable".}
\vspace{2mm}
\label{tab:effects_neg_sample}
\resizebox{0.8\linewidth}{!}{
\begin{tabular}{llccc}
\toprule
 \multirow{2}*{\textbf{Backbone}}& \multirow{2}*{\textbf{Model}} & \multicolumn{3}{c}{\textbf{Negative Sampling Strategy}} \\ 
 \cmidrule{3-5}
 & & \textbf{history} & \textbf{global} & \textbf{genertaed} \\

\midrule
\multirow{4}{*}{GNN} &
\textbf{ConvTransE}  
& 0.3737 & 0.5237 & N/A \\
& \textbf{RGCN}  & 0.3777 & 0.5272 & N/A \\
& \textbf{RE-GCN} & 0.3879 & 0.5407 & N/A \\
& \textbf{LoGo}  & \textbf{0.3969} & 0.5702 & N/A \\
\midrule
\multirow{2}{*}{gpt-3.5-turbo}
& \textbf{rule-based history} & 0.3290 & 0.6189 & 0.4632 \\
& \textbf{retrieved history} & 0.3533 & \textbf{0.6433} & \textbf{0.4751} \\
\bottomrule
\end{tabular}}
\end{table}

\subsection{Effects of RAG (RQ2)}
We conduct an investigation into the characteristics of retrieval in temporal event forecasting from multiple perspectives, including retrieval model, retrieval scope, and historical length.
\subsubsection{Performance \wrt Various Retrieval Models}
Different retrieval models may affect the performance. To study this problem, for text-involved settings, \ie Text-only and Mixed, we try three different retrieval models, including the classical BM25 and more recent methods of Contriver and LlamaIndex. From the results in Table~\ref{tab:rag_retrieval_model}, we can observe that the LlamaIndex > Contriver > BM25, in terms of forecasting accuracy on both settings. These results justify our hypothesis that a stronger retriever always yields better forecasting performance. This further shows that it is a promising direction to incorporate more powerful retrievers or even design event forecasting oriented retrievers. While for graph-only methods, the research on incorporating various graph retrieval models into LLMs is still under exploration, which is also expected to be extensively studied in the future. 
\begin{table}
\centering
\caption{Performance comparison of using different retrieval models on the Text-only and Mixed settings. }
\renewcommand{\arraystretch}{1.2}

\label{tab:rag_retrieval_model}
\resizebox{0.5\linewidth}{!}{
\begin{tabular}{lcc}
\toprule
 \textbf{Retriever}        & \textbf{Text-only} & \textbf{Mixed}\\
\midrule
\textbf{BM25~\citep{BM25}}  &0.2893 &0.3129 \\
\textbf{Contriver~\citep{Contriever}}  &0.3015 &0.3285 \\
\textbf{LlamaIndex~\citep{LlamaIndex}} &\textbf{0.3228} &\textbf{0.3539}\\
\bottomrule
\end{tabular}}
\end{table}

\begin{table}
\centering
\caption{Performance comparison \wrt applying different retrieval scopes. "N/A" means "Not Applicable". "$\ast $" indicates no historical input and it is falls into graph-only.}
\vspace{3mm}
\label{tab:rag_scope}
\resizebox{0.6\linewidth}{!}{
\begin{tabular}{l ccc}
\toprule
\shortstack[l]{\textbf{Model} \\ \textbf{(gpt-3.5-turbo)}} & \textbf{Graph-only}        & \textbf{Text-only} & \textbf{Mixed}\\
\midrule
\textbf{without history}  & $0.2746^*$ & N/A & N/A\\
\textbf{global history} &0.3403 &0.2429 &0.3267\\
\textbf{complex event} &\textbf{0.3533} &\textbf{0.3228} &\textbf{0.3539}\\
\bottomrule
\end{tabular}}
\end{table}
\subsubsection{Performance \wrt Various Retrieval Scopes}
To comprehensively examine the influence of retrieval scope on temporal event forecasting, we conduct a comparative analysis of the RAG model with three different retrieval options: \textit{without history} (\ie without using retrieval), \textit{global retrieval} (\ie retrieve from the global context that includes all complex events), and \textit{complex event} (\ie retrieve from within complex event). Upon analyzing the results shown in Table~\ref{tab:rag_scope}, we find out that regardless of whether the retrieval is performed on graphs, text, or a combination of both, the performance of complex event retrieval surpasses the other options. This can be primarily attributed to the fact that complex events provide a comprehensive depiction of the development and evolution of a sequence of events, offering rich contextual information. Thus, the retrieval scope is significantly narrowed down, enabling better performance. 
In contrast, \textit{global} retrieval introduces a considerable amount of noise during the retrieval process, leading to a decrease in retrieval efficiency compared to complex event retrieval. Nonetheless, it still provides some historical information, resulting in noticeable improvement when compared to the absence of retrieval. Moreover, these findings highlight the effectiveness of retrieval enhancement in large language models for predicting temporal events.

\begin{figure}
    \centering
    \includegraphics[width=0.8\linewidth]{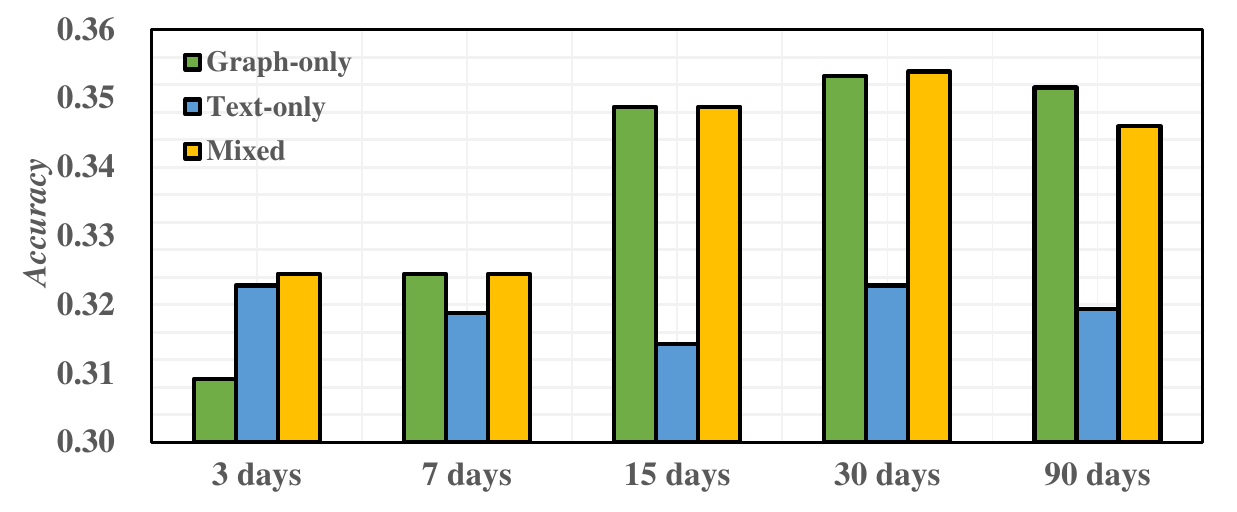}
    \caption{Performance comparison considering varying historical length for the model of "retrieved history".}
    \label{fig:rag_history_length}
\end{figure}

\subsubsection{Performance \wrt Varying Historical Length}
The scope of history can be interpreted as the spatial dimensionality of the world, while the temporal dimensionality does also matter. Especially in the task of temporal event forecasting, properly setting the historical length, even with the utilization of retrieval models, may play a crucial role in effective forecasting. Driven by this hypothesis, we test our default model of "retrieved history" in terms of different historical lengths, ranging from $\{3, 7, 15, 30, 90\}$ days. In other words, we constrain the retrieval models to search within the specified period of history. For example, three days of historical length means the retrieval models can only search from the events/articles in the past three days. We demonstrate the results in Figure~\ref{fig:rag_history_length}. For graph-involved methods (Graph-only and Mixed), the performance first grows and then drops, with the historical length increasing. This phenomenon makes sense because when the given history is very short, the useful information for sound forecasting is likely to be in the farther past. Hence, no matter how powerful the retrieval model is, it will be bottle-necked by the limited accessible history. When the historical length increases, more information is brought to the view, resulting in increased performance. However, the expansion of history length comes along with more noisy information, which is why when the historical length exceeds some threshold, the performance starts dropping. Given its poor performance, such a phenomenon does not appear in the Text-only setting. In this case, it may be because the text-based retrieval models constrain the effectiveness of the retrieval process, making the forecasting method insensitive to historical length.

\begin{figure}
    \centering
    \includegraphics[width=0.8\linewidth]{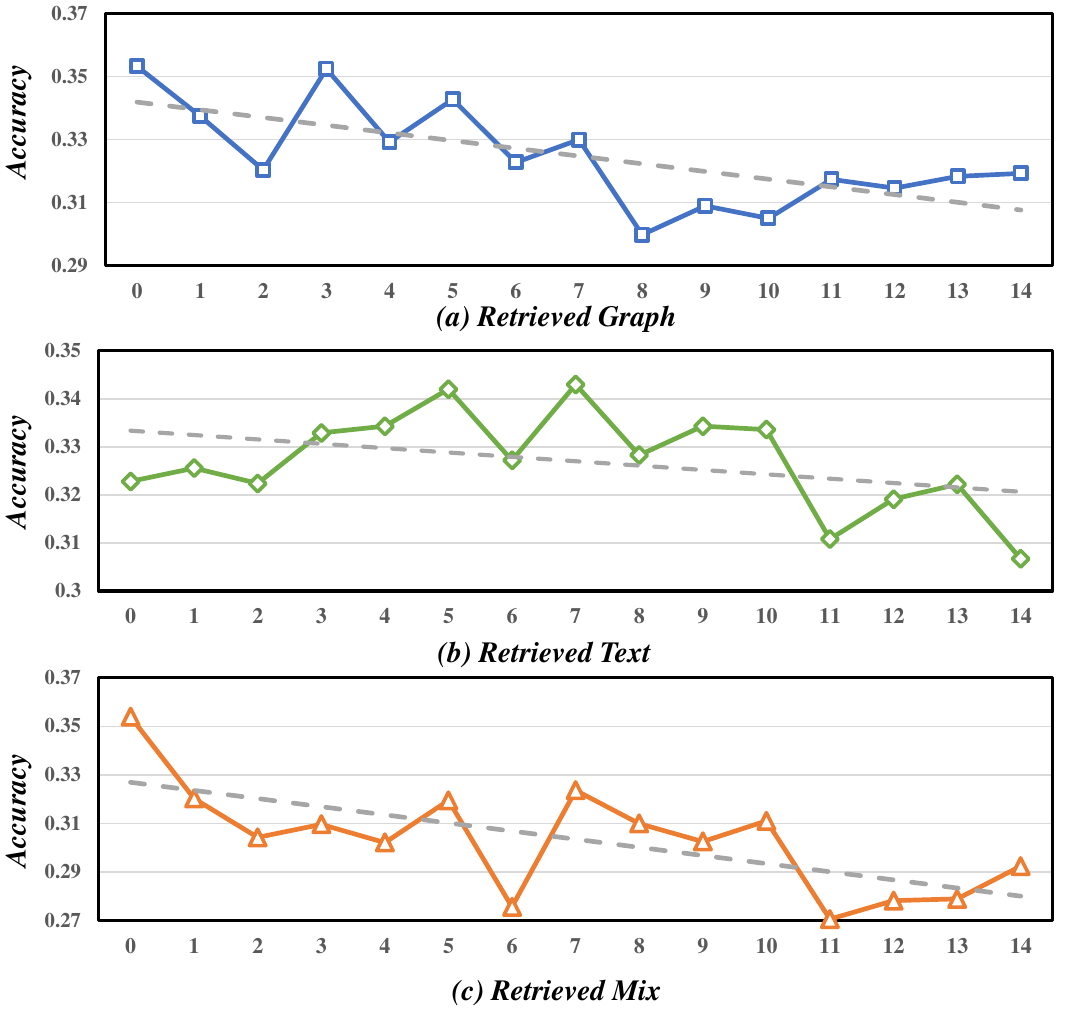}
    \caption{The performance comparison on long-term forecasting, where the sub-figure shows the accuracy of the different "retrieved history" models. The horizontal axis represents the time interval between the current timestamp and the query timestamp. }
    \label{fig:long_term_forecasting}
\end{figure}

\subsubsection{Performance \wrt Long-Term Forecasting}

In temporal event forecasting, forecasting events over longer time horizons can be more challenging due to increased uncertainty, the influence of external factors, and the potential for unforeseen events. To evaluate the performance of LLM-based methods in long-term forecasting, we tested our default "retrieved history" model to predict events occurring within a future time frame ranging from 1 to 14 days. For example, a time interval of 7 days signifies using historical events before timestamp t to predict events occurring at $t+7$. As shown in Figure~\ref{fig:long_term_forecasting}, the performance of LLM-based methods, as expected, deteriorates as the temporal gap between future and current timestamps increases. Furthermore, the variation in prediction accuracy across different time points may result from the uneven distribution of events at various stages of development. In other words, recent events may not be highly related to the query event, meaning that short-term changes in the query event might not significantly influence the prediction. Additionally, the performance on the mixed setting is worse than the graph-only and text-only settings. This may be because the mixed retrieval method provides more fine-grained contextual information but also introduces a considerable amount of irrelevant noise.

\begin{figure}
    \centering
    \includegraphics[width=0.7\linewidth]{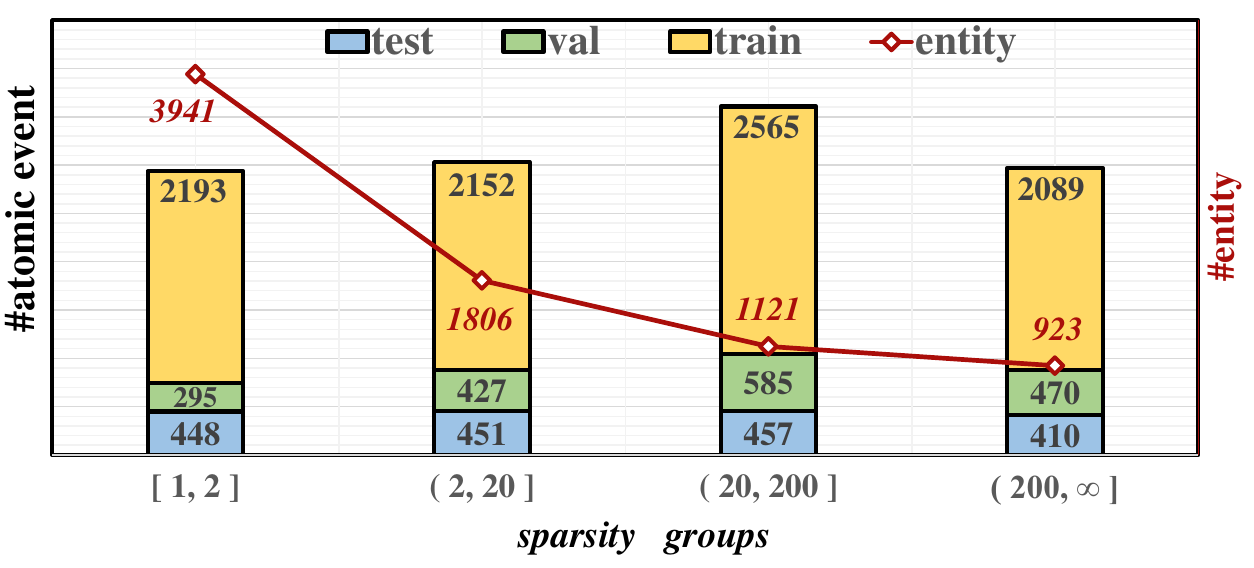}
    \caption{The statistics of different sparsity (density) groups, where bars show the number of atomic events and lines indicate the number of entities in each group.}
    \label{fig:pop_bias_stats}
\end{figure}

\subsection{Effects of Popularity Bias (RQ3)}
Several previous works, such as SeCoGD~\citep{SeCoGD}, have identified that event forecasting suffers from severe popularity bias. To study this interesting problem, we group the testing atomic events into four clusters according to the sparsity (popularity) degree of the object entity~\footnote{Alternative ways to measure the sparsity degree of an atomic event are using the occurrence frequency of subject entity or the average frequency of both subject and object. We leave these additional settings for future work.} within each quadruple. As shown in Figure~\ref{fig:pop_bias_stats}, each bar corresponds to one group of entities and its x-axis label denotes the occurrence frequency span. The height of the bar represents the number of atomic events whose object entities fall in the group, meanwhile, we separately annotate the number of atomic events in train/val/test sets for each group. The line denotes the number of entities (including both subject and object entities) in each group. Apparently, with a similar size of atomic events, the sparser groups have larger sets of entities while denser groups have fewer, exhibiting significant popularity bias (long-tail) phenomenon. 
\begin{figure}
    \centering
    \includegraphics[width=0.8\linewidth]{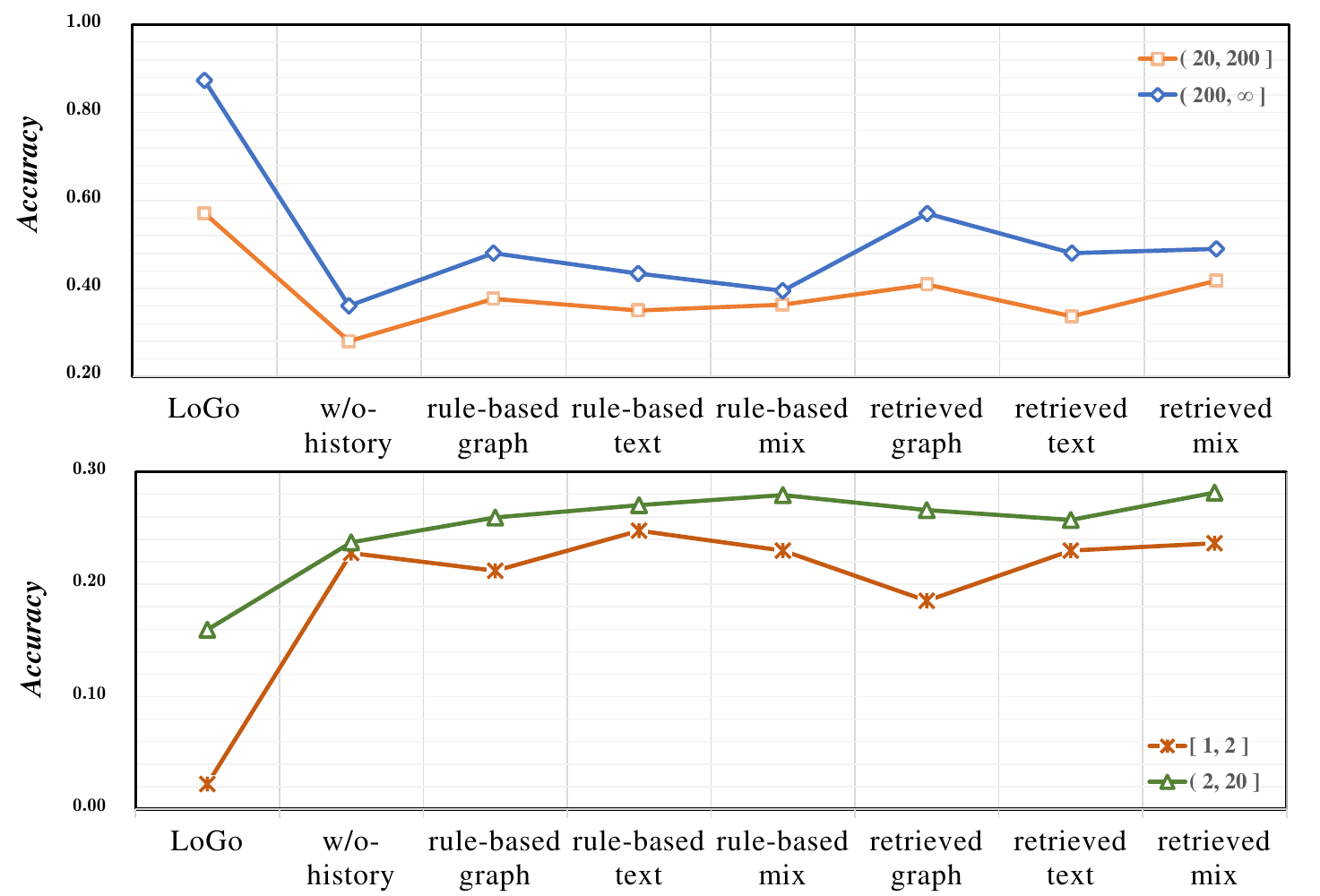}
    \caption{The breakdown performance comparison on different sparsity (density) groups, where the top sub-figure shows two denser (more popular) groups while the bottom sub-figure illustrates two sparser groups. The horizontal axis corresponds to multiple methods including both non-LLM method (LoGo) and LLM-based methods.}
    \label{fig:pop_bias_results}
\end{figure}
To study the popularity bias issue, we make statistics of the performance of several methods according to the sparsity groups, as illustrated in Figure~\ref{fig:pop_bias_results}. Each line corresponds to the performance for one sparsity group on various methods, where the top sub-figure presents the two denser groups while the bottom sub-figure shows the two sparser groups. Interestingly, we can observe that the non-LLM method, \ie LoGo, exhibits significant performance gaps between sparse and dense groups. In contrast, LLM-based methods, under no matter what specific settings, have much smaller gaps. This is because LoGo has been trained on the training set, thus simply fitting the entity distribution in the dataset. Thereafter, LoGo demonstrates the strongest overall performance; however, in long-tail sparse entity groups, LoGo performs much worse than LLM-based models. This presents that LLMs without fine-tuning are more robust to popularity bias, being able to generate better results for long-tail sparse entities. Among all the LLM-based methods, \textit{retrieved graph} demonstrates the largest performance gap between sparse and dense groups. This phenomenon raises a concern that graph-based retrieval would exacerbate the popularity bias issue since the retrieval modules are also affected by popularity bias.

\begin{figure}
    \centering
    \includegraphics[width=\linewidth]{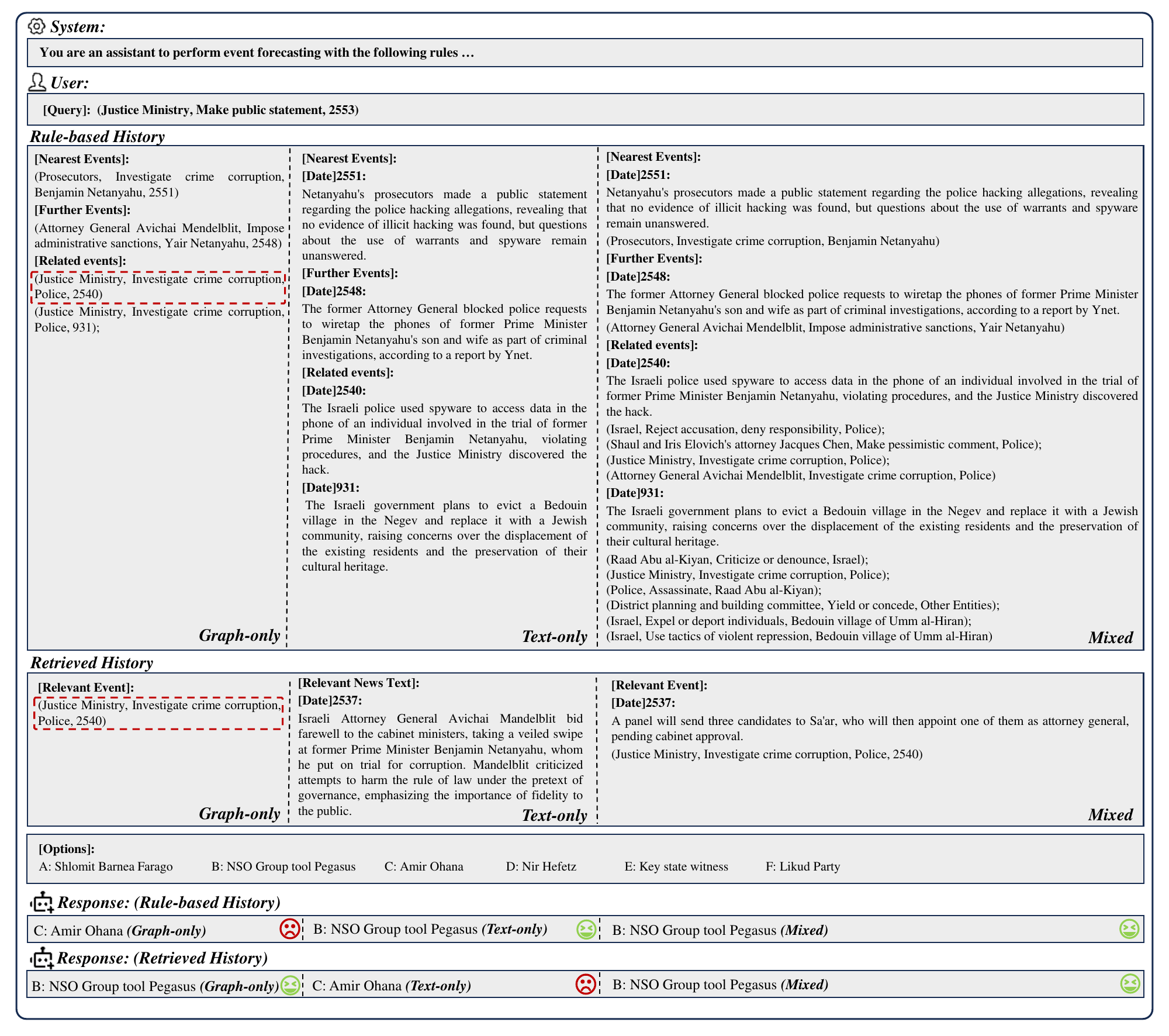}
    \caption{Case Study: Examples of our methods that consider all three input formats, encompassing both rule-based and retrieved history.}
    \label{fig:case_study}
\end{figure}

\subsection{Case Study}
In this section, we conduct a qualitative analysis of our methods \wrt various input formats and histories. The details of these cases are presented in Figure~\ref{fig:case_study}
, from which we can observe several interesting findings. First, graph-based history is more effective at modeling temporal and relational patterns, but it also amplifies the impact of noise. For example, in the graph-only setting, rule-based history includes multiple historical events that provide rich historical information but may interfere with the reasoning capabilities of LLMs. Conversely, retrieved history is constructed by extracting triples, such as (Justice Ministry, Investigate crime corruption, Police, 2540). The connection between "Police" and "NSO Group tool Pegasus" enables LLMs to infer the correct answer. Second, news texts can provide richer contextual information for event forecasting, thereby aiding the reasoning process of LLMs. However, the inclusion of historical news texts in the input is limited by token constraints and the ability of LLMs to handle long texts. Compared to rule-based history, retrieval-augmented generation (RAG) can significantly reduce the token length of the input while retaining the most relevant historical information. Nonetheless, retrieved news texts based on semantic similarity do not necessarily reflect the development trends of events, which can lead to incorrect answers in the text-only setting when using retrieved history. This highlights the importance of more powerful retrieval mechanisms to improve the performance of text-based methods.

\section{Related Work}
We briefly survey the related works from two perspectives: general works on temporal event forecasting and more specifically, large language model-based methods in temporal event forecasting.
\subsection{Temporal Event Forecasting}
Temporal event forecasting aims to predict future events based on the observed historical events. Different formulations have been raised for this task~\citep{survey-event-forecasting}. 
They can generally be grouped into three categories based on the event format: time series, unstructured textual events, and structured events. 

Some works model event occurrence or features using time series~\citep{timeSeriesEvent,benjamin2023hybrid,RCT_B}, but they fail to represent multi-relations among entities and multi-line natural of events. 
Several studies also explore the unstructured textual representation of temporal events, where each atomic event is generated from multi-documents in the form of summary~\citep{gholipour-ghalandari-etal-2020-large} or phrases~\citep{eventChain} and is chained in temporal order. 
Several studies have also investigated unstructured textual representations of temporal events, in which each atomic event is derived from multiple documents and expressed either as a summary\citep{gholipour-ghalandari-etal-2020-large}, and then linked sequentially in temporal order.
The natural language event description contains more fine-grained details but also leads to higher information complexity that impedes the downstream forecasting task formulation and performance. ForecastQA~\citep{ForecastQA} formulates the event forecasting task in a question-answering manner with a retrieval database consisting of all documents that contain historical events. However, due to the lack of structure extracted events, the generation of QA pairs requires heavy human labeling and extensive domain knowledge.

Efforts have also been exhausted in structured event forecasting. 
The major line of work represents temporal events using temporal knowledge graphs where each atomic event is a time-stamped link~\citep{GDELT, ICEWS}. Extensive methods have been proposed for event forecasting, including approaches that integrate temporal and relational dependencies among entities~\citep{RENET, REGCN, EvoKG}, retrieve relevant historical events~\citep{cygnet, TimeTraveler, cluster}, or model the continuous temporal evolution of events~\citep{knowevolve, TANGO}. RGCN~\citep{RGCN} adopts a static knowledge graph to capture the relational patterns among entities. RENET~\citep{RENET} utilizes RGCN and the gated recurrent units (GRU) to model the evolution of the temporal knowledge graph over time for event forecasting. REGCN ~\citep{REGCN} adopts ConvTransE~\citep{ConvTransE} as the decoder to predict the event at the next time stamp. Some methods have also tried to incorporate textual event information into TKG. 
Considering the absence of specific and informative events,SeCoGD~\citep{SeCoGD} employs textual topic modeling to disentangle subgraphs, leveraging the MidEast-TE dataset for event forecasting. In contrast, LoGo~\citep{LoGo} harnesses text clustering to construct complex events for forecasting, incorporating both local and global contexts.
In order to incorporate textual information into representation learning, Glean~\citep{Glean}  and CMF~\citep{CMF} fuse textual embeddings in graph edges. However, all of them still conduct the forecast reasoning on graphs, while in our work, we investigate forecasting in a hybrid setting of leveraging both text and graph.

\subsection{LLMs in Temporal Event Forecasting}
Generative Language Models (LM), especially the LLMs~\citep{GPT4}, have shown superior capability in language understanding and reasoning across a wide range of tasks and domains, including science\citep{PoT, scibench, mathvista} and healthcare~\citep{PMCLLaMA, medicineLLM}. Researchers have also conducted various studies on LLMs for temporal reasoning. One line of work focuses on temporal understanding where LLMs are tested for temporal event ordering or storyline understanding~\citep{tempbench, torque, going, SentimentAnalysis, tram}. Compared to the understanding task, the forecasting task is generally of higher difficulty where the prediction targets do not appear in the input and, therefore, require the model to conduct necessary inferences. Recent work has employed Large Language Models (LLMs) for temporal event forecasting by transforming Temporal Knowledge Graph (TKG) formulations into textual sequences, and recasting missing object prediction as a next-token prediction task~\citep{PPT}. For instance, GPT-NeoX-ICL\citep{GPT-NeoX-ICL} utilizes in-context learning by structuring prompts as a list of historical events, each represented in quadruplet format. Meanwhile, LAMP\citep{LMAP} applies LLM-based abductive reasoning to enhance retrieval in forecasting pipelines.
GENTKG~\citep{GENTKG} learns temporal logical rules via temporal random walks and retrieves historical events by these rules to help LLMs reasoning on the TKG. However, all of these works only investigate LLMs on temporal event forecasting with structured graph event data. The evaluation of LLMs on textual temporal event forecasting remains explored, not to mention the hybrid setting of graph and text that we also study in this work.

\section{Conclusion and Future Work}

In this paper, we systematically evaluated LLM-based methods on the task of text-involved temporal event forecasting. Specifically, we first built a benchmark dataset by using the SOTA LLM GPT-4 as the event extractor. Then we designed a series of LLM-based event forecasting models equipped with multiple configurable components, including optional input modalities, different forecasting objectives, different backbone LLMs, fine-tuning, and RAG with various settings. We also studied how popularity bias affects temporal event forecasting. Implementing all these model variations, we obtained a comprehensive understanding of the current status of how LLMs perform on text-involved temporal event forecasting. More importantly, from the extensive evaluation, we pinpointed several key research questions that are meaningful while challenging. First, developing effective text-based retrieval models in the context of temporal event forecasting is essentially important, given the large amount unstructured news articles posted online every day. Second, popularity bias and long-tail issues are still severe, and pertinent measurement and mitigation approaches are highly desired. Finally, developing event forecasting-oriented LLMs, \ie tuning a task-specific LLM, seems to be the most promising direction.

Moving forward, to address the identified key challenges, there is a lot of work to be done, and here we would like to highlight two aspects. First, building larger, more accurate, and versatile benchmark datasets is pressing. Event forecasting is featured with highly domain-specific characteristics, therefore, it is hard to obtain a generalizable conclusion on small-scaled noisy datasets in one or a small number of domains, which is also the major limitation of this work. Second, integrating structural and textual information into the reasoning process of LLMs is crucial for improving the performance of temporal event forecasting. While fine-tuning and retrieval-augmented generation (RAG) show promise, they remain constrained by token length limitations and the inherent gap between structured graph data and unstructured textual data. Thus, exploring the alignment of graph representations with textual representations within the token embedding space of LLMs is a promising direction for future research. Finally, focusing on task-level characteristics instead of arms racing on overall performance is necessary and more valuable. Event forecasting is a long-standing but slow-developing research area, one non-negligible reason of which is under exploration and poor understanding of problem settings.

\section*{CRediT authorship contribution statement}
\textbf{He Chang}: Conceptualization, Methodology, Validation, Writing-Original Draft. \textbf{Chenchen Ye}: Conceptualization, Investigation, Writing- Review \& Editing. \textbf{Zhulin Tao}: Writing-Review \& Editing, Project administration. Jie Wu: Investigation, DataCuration. \textbf{Zhengmao Yang}: Formalanalysis, DataCuration. \textbf{Yunshan Ma}: Conceptualization of this study, Resources, Writing- Original Draft, Supervision. \textbf{Xianglin Huang}: Resources, Writing-Review \& Editing, Project administration. \textbf{Tat-Seng Chua}: Resources, Writing- Review \& Editing, Project administration.








\bibliographystyle{elsarticle-num}
\bibliography{main}

\begin{thebibliography}{10}
\expandafter\ifx\csname url\endcsname\relax
  \def\url#1{\texttt{#1}}\fi
\expandafter\ifx\csname urlprefix\endcsname\relax\def\urlprefix{URL }\fi
\expandafter\ifx\csname href\endcsname\relax
  \def\href#1#2{#2} \def\path#1{#1}\fi

\bibitem{survey-event-forecasting}
L.~Zhao, Event prediction in the big data era: {A} systematic survey, {ACM} Comput. Surv. 54~(5) (2022) 94:1--94:37.

\bibitem{GDELT}
K.~Leetaru, P.~A. Schrodt, Gdelt: Global data on events, location, and tone, 1979--2012, in: ISA annual convention, Vol.~2, Citeseer, 2013, pp. 1--49.

\bibitem{ICEWS}
S.~P. O'brien, Crisis early warning and decision support: Contemporary approaches and thoughts on future research, International studies review 12~(1) (2010) 87--104.

\bibitem{Glean}
S.~Deng, H.~Rangwala, Y.~Ning, Dynamic knowledge graph based multi-event forecasting, in: {KDD}, {ACM}, 2020, pp. 1585--1595.

\bibitem{TKG-survey}
B.~Cai, Y.~Xiang, L.~Gao, H.~Zhang, Y.~Li, J.~Li, Temporal knowledge graph completion: {A} survey, CoRR abs/2201.08236 (2022).

\bibitem{LoGo}
Y.~Ma, C.~Ye, Z.~Wu, X.~Wang, Y.~Cao, L.~Pang, T.~Chua, Structured, complex and time-complete temporal event forecasting, CoRR abs/2312.01052 (2023).

\bibitem{ConvE}
T.~Dettmers, P.~Minervini, P.~Stenetorp, S.~Riedel, Convolutional 2d knowledge graph embeddings, in: {AAAI}, {AAAI} Press, 2018, pp. 1811--1818.

\bibitem{ConvTransE}
C.~Shang, Y.~Tang, J.~Huang, J.~Bi, X.~He, B.~Zhou, End-to-end structure-aware convolutional networks for knowledge base completion, in: {AAAI}, {AAAI} Press, 2019, pp. 3060--3067.

\bibitem{RGCN}
M.~S. Schlichtkrull, T.~N. Kipf, P.~Bloem, R.~van~den Berg, I.~Titov, M.~Welling, Modeling relational data with graph convolutional networks, in: {ESWC}, Vol. 10843 of Lecture Notes in Computer Science, Springer, 2018, pp. 593--607.

\bibitem{theFuture}
M.~Li, S.~Li, Z.~Wang, L.~Huang, K.~Cho, H.~Ji, J.~Han, C.~R. Voss, The future is not one-dimensional: Complex event schema induction by graph modeling for event prediction, in: {EMNLP} {(1)}, Association for Computational Linguistics, 2021, pp. 5203--5215.

\bibitem{SeCoGD}
Y.~Ma, C.~Ye, Z.~Wu, X.~Wang, Y.~Cao, T.~Chua, Context-aware event forecasting via graph disentanglement, in: {KDD}, {ACM}, 2023, pp. 1643--1652.

\bibitem{RENET}
W.~Jin, M.~Qu, X.~Jin, X.~Ren, Recurrent event network: Autoregressive structure inferenceover temporal knowledge graphs, in: {EMNLP} {(1)}, Association for Computational Linguistics, 2020, pp. 6669--6683.

\bibitem{REGCN}
Z.~Li, X.~Jin, W.~Li, S.~Guan, J.~Guo, H.~Shen, Y.~Wang, X.~Cheng, Temporal knowledge graph reasoning based on evolutional representation learning, in: {SIGIR}, {ACM}, 2021, pp. 408--417.

\bibitem{EvoKG}
N.~Park, F.~Liu, P.~Mehta, D.~Cristofor, C.~Faloutsos, Y.~Dong, Evokg: Jointly modeling event time and network structure for reasoning over temporal knowledge graphs, in: {WSDM}, {ACM}, 2022, pp. 794--803.

\bibitem{MA2024103848}
J.~Ma, K.~Li, F.~Zhang, Y.~Wang, X.~Luo, C.~Li, Y.~Qiao, Taret: Temporal knowledge graph reasoning based on topology-aware dynamic relation graph and temporal fusion, Information Processing \& Management 61~(6) (2024) 103848.
\newblock \href {https://doi.org/https://doi.org/10.1016/j.ipm.2024.103848} {\path{doi:https://doi.org/10.1016/j.ipm.2024.103848}}.

\bibitem{scriptLearning}
S.~Lv, F.~Zhu, S.~Hu, Integrating external event knowledge for script learning, in: Proceedings of the 28th International Conference on Computational Linguistics, 2020, pp. 306--315.

\bibitem{ForecastQA}
W.~Jin, R.~Khanna, S.~Kim, D.~Lee, F.~Morstatter, A.~Galstyan, X.~Ren, Forecastqa: {A} question answering challenge for event forecasting with temporal text data, in: {ACL/IJCNLP} {(1)}, Association for Computational Linguistics, 2021, pp. 4636--4650.

\bibitem{CMF}
S.~Deng, H.~Rangwala, Y.~Ning, Understanding event predictions via contextualized multilevel feature learning, in: {CIKM}, {ACM}, 2021, pp. 342--351.

\bibitem{GPT}
T.~Kojima, S.~S. Gu, M.~Reid, Y.~Matsuo, Y.~Iwasawa, Large language models are zero-shot reasoners, in: NeurIPS, 2022.

\bibitem{RLHF}
L.~Ouyang, J.~Wu, X.~Jiang, D.~Almeida, C.~L. Wainwright, P.~Mishkin, C.~Zhang, S.~Agarwal, K.~Slama, A.~Ray, J.~Schulman, J.~Hilton, F.~Kelton, L.~Miller, M.~Simens, A.~Askell, P.~Welinder, P.~F. Christiano, J.~Leike, R.~Lowe, Training language models to follow instructions with human feedback, in: NeurIPS, 2022.

\bibitem{LLaMA}
H.~Touvron, T.~Lavril, G.~Izacard, X.~Martinet, M.~Lachaux, T.~Lacroix, B.~Rozi{\`{e}}re, N.~Goyal, E.~Hambro, F.~Azhar, A.~Rodriguez, A.~Joulin, E.~Grave, G.~Lample, Llama: Open and efficient foundation language models, CoRR abs/2302.13971 (2023).

\bibitem{PPT}
W.~Xu, B.~Liu, M.~Peng, X.~Jia, M.~Peng, Pre-trained language model with prompts for temporal knowledge graph completion, in: {ACL} (Findings), Association for Computational Linguistics, 2023, pp. 7790--7803.

\bibitem{GPT-NeoX-ICL}
D.~Lee, K.~Ahrabian, W.~Jin, F.~Morstatter, J.~Pujara, Temporal knowledge graph forecasting without knowledge using in-context learning, in: {EMNLP}, Association for Computational Linguistics, 2023, pp. 544--557.

\bibitem{GENTKG}
R.~Liao, X.~Jia, Y.~Ma, V.~Tresp, Gentkg: Generative forecasting on temporal knowledge graph, CoRR abs/2310.07793 (2023).

\bibitem{Chain-of-History}
R.~Luo, T.~Gu, H.~Li, J.~Li, Z.~Lin, J.~Li, Y.~Yang, Chain of history: Learning and forecasting with llms for temporal knowledge graph completion, CoRR abs/2401.06072 (2024).

\bibitem{ToG}
J.~Sun, C.~Xu, L.~Tang, S.~Wang, C.~Lin, Y.~Gong, L.~M. Ni, H.~Shum, J.~Guo, Think-on-graph: Deep and responsible reasoning of large language model on knowledge graph, in: {ICLR}, OpenReview.net, 2024.

\bibitem{liang2024synergizing}
J.~Liang, L.~Liao, H.~Fei, J.~Jiang, Synergizing large language models and pre-trained smaller models for conversational intent discovery, in: Findings of the Association for Computational Linguistics ACL 2024, 2024, pp. 14133--14147.

\bibitem{RAG}
P.~S.~H. Lewis, E.~Perez, A.~Piktus, F.~Petroni, V.~Karpukhin, N.~Goyal, H.~K{\"{u}}ttler, M.~Lewis, W.~Yih, T.~Rockt{\"{a}}schel, S.~Riedel, D.~Kiela, Retrieval-augmented generation for knowledge-intensive {NLP} tasks, in: NeurIPS, 2020.

\bibitem{sun-etal-2023-chatgpt}
W.~Sun, L.~Yan, X.~Ma, S.~Wang, P.~Ren, Z.~Chen, D.~Yin, Z.~Ren, Is {C}hat{GPT} good at search? investigating large language models as re-ranking agents, in: Proceedings of the 2023 Conference on Empirical Methods in Natural Language Processing, Association for Computational Linguistics, 2023, pp. 14918--14937.
\newblock \href {https://doi.org/10.18653/v1/2023.emnlp-main.923} {\path{doi:10.18653/v1/2023.emnlp-main.923}}.

\bibitem{GRLC}
L.~Peng, Y.~Mo, J.~Xu, J.~Shen, X.~Shi, X.~Li, H.~T. Shen, X.~Zhu, Grlc: Graph representation learning with constraints, IEEE Transactions on Neural Networks and Learning Systems 35~(6) (2024) 8609--8622.
\newblock \href {https://doi.org/10.1109/TNNLS.2022.3230979} {\path{doi:10.1109/TNNLS.2022.3230979}}.

\bibitem{CAMEO}
E.~Boschee, J.~Lautenschlager, S.~O'Brien, S.~Shellman, J.~Starz, M.~Ward, \href{https://doi.org/10.7910/DVN/28075/SCJPXX}{Cameo.cdb.09b5.pdf}, in: ICEWS Coded Event Data, Harvard Dataverse, 2015.
\newblock \href {https://doi.org/10.7910/DVN/28075/SCJPXX} {\path{doi:10.7910/DVN/28075/SCJPXX}}.
\newline\urlprefix\url{https://doi.org/10.7910/DVN/28075/SCJPXX}

\bibitem{vicuna2023}
W.-L. Chiang, Z.~Li, Z.~Lin, Y.~Sheng, Z.~Wu, H.~Zhang, L.~Zheng, S.~Zhuang, Y.~Zhuang, J.~E. Gonzalez, I.~Stoica, E.~P. Xing, \href{https://lmsys.org/blog/2023-03-30-vicuna/}{Vicuna: An open-source chatbot impressing gpt-4 with 90\%* chatgpt quality} (March 2023).
\newline\urlprefix\url{https://lmsys.org/blog/2023-03-30-vicuna/}

\bibitem{QLoRA}
T.~Dettmers, A.~Pagnoni, A.~Holtzman, L.~Zettlemoyer, Qlora: Efficient finetuning of quantized llms, CoRR abs/2305.14314 (2023).

\bibitem{BM25}
S.~Robertson, H.~Zaragoza, et~al., The probabilistic relevance framework: Bm25 and beyond, Foundations and Trends{\textregistered} in Information Retrieval 3~(4) (2009) 333--389.

\bibitem{Contriever}
G.~Izacard, M.~Caron, L.~Hosseini, S.~Riedel, P.~Bojanowski, A.~Joulin, E.~Grave, Unsupervised dense information retrieval with contrastive learning, arXiv preprint arXiv:2112.09118 (2021).

\bibitem{LlamaIndex}
J.~Liu, {LlamaIndex} (11 2022).
\newblock \href {https://doi.org/10.5281/zenodo.1234} {\path{doi:10.5281/zenodo.1234}}.

\bibitem{timeSeriesEvent}
N.~B. Weidmann, M.~D. Ward, Predicting conflict in space and time, Journal of Conflict Resolution 54~(6) (2010) 883--901.

\bibitem{benjamin2023hybrid}
D.~M. Benjamin, F.~Morstatter, A.~E. Abbas, A.~Abeliuk, P.~Atanasov, S.~Bennett, A.~Beger, S.~Birari, D.~V. Budescu, M.~Catasta, et~al., Hybrid forecasting of geopolitical events, AI Magazine (2023).

\bibitem{RCT_B}
F.~Morstatter, \href{https://doi.org/10.7910/DVN/ROTHFT}{{RCT-B}} (2021).
\newblock \href {https://doi.org/10.7910/DVN/ROTHFT} {\path{doi:10.7910/DVN/ROTHFT}}.
\newline\urlprefix\url{https://doi.org/10.7910/DVN/ROTHFT}

\bibitem{gholipour-ghalandari-etal-2020-large}
D.~Gholipour~Ghalandari, C.~Hokamp, N.~T. Pham, J.~Glover, G.~Ifrim, \href{https://aclanthology.org/2020.acl-main.120}{A large-scale multi-document summarization dataset from the {W}ikipedia current events portal}, in: D.~Jurafsky, J.~Chai, N.~Schluter, J.~Tetreault (Eds.), Proceedings of the 58th Annual Meeting of the Association for Computational Linguistics, Association for Computational Linguistics, Online, 2020, pp. 1302--1308.
\newblock \href {https://doi.org/10.18653/v1/2020.acl-main.120} {\path{doi:10.18653/v1/2020.acl-main.120}}.
\newline\urlprefix\url{https://aclanthology.org/2020.acl-main.120}

\bibitem{eventChain}
Y.~Jiao, M.~Zhong, J.~Shen, Y.~Zhang, C.~Zhang, J.~Han, Unsupervised event chain mining from multiple documents, in: {WWW}, {ACM}, 2023, pp. 1948--1959.

\bibitem{cygnet}
C.~Zhu, M.~Chen, C.~Fan, G.~Cheng, Y.~Zhan, \href{https://api.semanticscholar.org/CorpusID:229180723}{Learning from history: Modeling temporal knowledge graphs with sequential copy-generation networks}, in: AAAI Conference on Artificial Intelligence, 2020.
\newline\urlprefix\url{https://api.semanticscholar.org/CorpusID:229180723}

\bibitem{TimeTraveler}
H.~Sun, J.~Zhong, Y.~Ma, Z.~Han, K.~He, \href{https://api.semanticscholar.org/CorpusID:237454564}{Timetraveler: Reinforcement learning for temporal knowledge graph forecasting}, in: EMNLP, 2021.
\newline\urlprefix\url{https://api.semanticscholar.org/CorpusID:237454564}

\bibitem{cluster}
Z.~Li, X.~Jin, S.~Guan, W.~Li, J.~Guo, Y.~Wang, X.~Cheng, \href{https://api.semanticscholar.org/CorpusID:235266233}{Search from history and reason for future: Two-stage reasoning on temporal knowledge graphs}, in: ACL, 2021.
\newline\urlprefix\url{https://api.semanticscholar.org/CorpusID:235266233}

\bibitem{knowevolve}
R.~Trivedi, H.~Dai, Y.~Wang, L.~Song, Know-evolve: deep temporal reasoning for dynamic knowledge graphs, in: ICML, 2017, p. 3462–3471.

\bibitem{TANGO}
Z.~Ding, Z.~Han, Y.~Ma, V.~Tresp, \href{https://api.semanticscholar.org/CorpusID:231592393}{Temporal knowledge graph forecasting with neural ode} abs/2101.05151 (2021).
\newline\urlprefix\url{https://api.semanticscholar.org/CorpusID:231592393}

\bibitem{GPT4}
OpenAI, {GPT-4} technical report, CoRR abs/2303.08774 (2023).

\bibitem{PoT}
W.~Chen, X.~Ma, X.~Wang, W.~W. Cohen, Program of thoughts prompting: Disentangling computation from reasoning for numerical reasoning tasks, in: {TMLR}, 2023.

\bibitem{scibench}
X.~Wang, Z.~Hu, P.~Lu, Y.~Zhu, J.~Zhang, S.~Subramaniam, A.~R. Loomba, S.~Zhang, Y.~Sun, W.~Wang, Scibench: Evaluating college-level scientific problem-solving abilities of large language models (2023).
\newblock \href {http://arxiv.org/abs/2307.10635} {\path{arXiv:2307.10635}}.

\bibitem{mathvista}
P.~Lu, H.~Bansal, T.~Xia, J.~Liu, C.~Li, H.~Hajishirzi, H.~Cheng, K.-W. Chang, M.~Galley, J.~Gao, Mathvista: Evaluating mathematical reasoning of foundation models in visual contexts, in: {ICLR}, 2024.

\bibitem{PMCLLaMA}
C.~Wu, X.~Zhang, Y.~Zhang, Y.~Wang, W.~Xie, \href{https://api.semanticscholar.org/CorpusID:263888272}{Pmc-llama: Further finetuning llama on medical papers}, ArXiv abs/2304.14454 (2023).
\newline\urlprefix\url{https://api.semanticscholar.org/CorpusID:263888272}

\bibitem{medicineLLM}
A.~J. Thirunavukarasu, D.~S.~J. Ting, K.~Elangovan, L.~Gutierrez, T.~F. Tan, D.~S.~W. Ting, Large language models in medicine, Vol.~29, 2023, p. 1930–1940.

\bibitem{tempbench}
Q.~Tan, H.~T. Ng, L.~Bing, \href{https://aclanthology.org/2023.acl-long.828}{Towards benchmarking and improving the temporal reasoning capability of large language models}, in: {ACL}, Association for Computational Linguistics, 2023, pp. 14820--14835.
\newblock \href {https://doi.org/10.18653/v1/2023.acl-long.828} {\path{doi:10.18653/v1/2023.acl-long.828}}.
\newline\urlprefix\url{https://aclanthology.org/2023.acl-long.828}

\bibitem{torque}
Q.~Ning, H.~Wu, R.~Han, N.~Peng, M.~Gardner, D.~Roth, \href{https://aclanthology.org/2020.emnlp-main.88}{{TORQUE}: A reading comprehension dataset of temporal ordering questions}, in: {EMNLP}, 2020, pp. 1158--1172.
\newblock \href {https://doi.org/10.18653/v1/2020.emnlp-main.88} {\path{doi:10.18653/v1/2020.emnlp-main.88}}.
\newline\urlprefix\url{https://aclanthology.org/2020.emnlp-main.88}

\bibitem{going}
B.~Zhou, D.~Khashabi, Q.~Ning, D.~Roth, \href{https://aclanthology.org/D19-1332}{{``}going on a vacation{''} takes longer than {``}going for a walk{''}: A study of temporal commonsense understanding}, in: {EMNLP}, 2019, pp. 3363--3369.
\newblock \href {https://doi.org/10.18653/v1/D19-1332} {\path{doi:10.18653/v1/D19-1332}}.
\newline\urlprefix\url{https://aclanthology.org/D19-1332}

\bibitem{SentimentAnalysis}
T.~Sun, L.~Jing, Y.~Wei, X.~Song, Z.~Cheng, L.~Nie, Dual consistency-enhanced semi-supervised sentiment analysis towards covid-19 tweets, IEEE Transactions on Knowledge and Data Engineering 35~(12) (2023) 12605--12617.
\newblock \href {https://doi.org/10.1109/TKDE.2023.3270940} {\path{doi:10.1109/TKDE.2023.3270940}}.

\bibitem{tram}
Y.~Wang, Y.~Zhao, Tram: Benchmarking temporal reasoning for large language models (2023).
\newblock \href {http://arxiv.org/abs/2310.00835} {\path{arXiv:2310.00835}}.

\bibitem{LMAP}
X.~Shi, S.~Xue, K.~Wang, F.~Zhou, J.~Y. Zhang, J.~ZHOU, C.~Tan, H.~Mei, Language models can improve event prediction by few-shot abductive reasoning, in: {NeurIPS}, 2023.

\end{thebibliography}
\end{document}